\documentclass[10pt,twocolumn,letterpaper]{article}
\usepackage[pagenumbers]{cvpr}
\usepackage[utf8]{inputenc}
\usepackage[T1]{fontenc}
\usepackage{array,tabularx}
\usepackage{microtype}
\usepackage[normalem]{ulem}
\usepackage{upquote}
\DeclareUnicodeCharacter{2191}{\ensuremath{\uparrow}}
\DeclareUnicodeCharacter{2212}{\ensuremath{-}}
\DeclareUnicodeCharacter{00D7}{\ensuremath{\times}}

\definecolor{cvprblue}{rgb}{0.21,0.49,0.74}
\usepackage[pagebackref,breaklinks,colorlinks,allcolors=cvprblue]{hyperref}
\def\confName{CVPR}
\def\confYear{2027}
\date{}

\author{%
Tianyi Hu\thanks{These three authors contributed equally as co-first authors.}\quad
Yangdong Li\footnotemark[1]\quad
Shuyan Liu\footnotemark[1]\\[0.4em]
Zihan Cao\thanks{These six authors contributed equally as co-second authors.}\quad
Jingze Liang\footnotemark[2]\quad
Shaojie Wu\footnotemark[2]\quad
Chenyang Feng\footnotemark[2]\quad
Jincheng Di\footnotemark[2]\quad
Weijie Xiao\footnotemark[2]\\[0.4em]
Liangjian Deng\thanks{Corresponding author: \href{mailto:liangjian.deng@uestc.edu.cn}{\texttt{liangjian.deng@uestc.edu.cn}}.}\\[0.6em]
University of Electronic Science and Technology of China\\
Chengdu, China
}

\title{RAWPrism: A Three-Task Dataset and a Unified Model for RAW-Domain Image Fusion}

\hypersetup{
  pdftitle={RAWPrism: A Three-Task Dataset and a Unified Model for RAW-Domain Image Fusion},
  pdfauthor={Tianyi Hu, Yangdong Li, Shuyan Liu, Zihan Cao, Jingze Liang, Shaojie Wu, Chenyang Feng, Jincheng Di, Weijie Xiao, Liangjian Deng},
  pdfsubject={Computer Vision and Pattern Recognition},
  pdfkeywords={RAW image fusion, multi-exposure fusion, multi-focus fusion, infrared-visible fusion}
}
\begin{document}
\maketitle
\begin{abstract}
Research on multi-exposure fusion (MEF), multi-focus fusion (MFF), and infrared--visible image fusion (IVIF) has largely focused on rendered RGB images, while unified fusion across these tasks in the RAW domain before image signal processing (ISP) remains relatively underexplored. To support research in this setting, we introduce the RAWPrism dataset, covering all three tasks, and the RAWPrism model, a lightweight unified multi-task fusion network. The RAWPrism dataset comprises 422 usable MEF scenes and 150 registered MFF pairs from our RAW captures, together with 1,500 FMB-derived IVIF samples, in which the visible RAW inputs are reconstructed by ReRAW and their RGB counterparts rendered by CycleISP. Scene-level partitions and task-specific rendering and scoring protocols connect these resources to method development and evaluation. To our knowledge, the RAWPrism model is the first to unify MEF, MFF, and IVIF directly in the RAW domain. It combines a CFA-aware representation, task-conditioned source gating, and a shared backbone to support all three tasks with one 3.58M-parameter checkpoint. Together, the dataset and model support the study of shared fusion mechanisms across exposure, focus, and spectral complementarity.
\end{abstract}

% Condensed introduction; detailed construction and protocol definitions follow in Sections 3–5.
\section{Introduction}
\label{sec:1}

Multi-exposure fusion (MEF), multi-focus fusion (MFF), and infrared--visible fusion (IVIF) combine observations that reveal different scene information. Exposure changes bright- and dark-region detail, focus changes local sharpness, and visible and thermal sensing reveal different aspects of scene appearance. Each task therefore requires spatially varying decisions about what to retain from two sources.

Unified RGB fusion methods learn these decisions within shared architectures \cite{ref20,ref21,ref22,ref23,ref24}. Their inputs have already undergone demosaicing, denoising, and nonlinear intensity processing \cite{ref12,ref18}. Captured RAW instead preserves the color filter array (CFA) samples. After black-level correction, unsaturated measurements retain approximately linear radiometric relationships and signal-dependent noise \cite{ref12,ref19}. These properties support exposure- and noise-aware combination before tone mapping, and focus selection before independent demosaicing. Bayer-frame merging \cite{ref18}, Bayer-domain focus fusion \cite{ref35}, and CFA-aware preprocessing \cite{ref36} motivate this approach, but do not jointly address all three tasks.

We construct the RAWPrism dataset to study this shared fusion problem. It provides 422 usable captured MEF scenes and 150 captured, registered MFF pairs. For IVIF, we extend 1,500 FMB scenes \cite{ref1} through a reconstruction route. Learned inverse ISPs enable RAW representations to be recovered from RGB \cite{ref13,ref14,ref37}: ReRAW reconstructs the visible RAW, and CycleISP \cite{ref38} renders it into a corresponding RGB input. Both representations share the same thermal observation and scene partition. Reconstructed IVIF inputs provide a complementary research setting; they are distinguished from captured sensor measurements throughout the paper.

The RAWPrism model learns source selection through a CFA-aware latent, a task-conditioned gate, and a shared backbone. Task-specific heads reconstruct the fused RAW. This architecture serves all three tasks with one 3.58M-parameter checkpoint. To our knowledge, no prior method unifies MEF, MFF, and IVIF directly in the RAW domain.

Our contributions are threefold. First, the dataset brings exposure acquisition, focus acquisition and registration, and visible RAW reconstruction into one three-task resource. Second, its scene-level partitions and explicit input, rendering, spatial-processing, and scoring protocols define the conditions for RAW/RGB comparisons. Third, the model provides a compact unified architecture for studying shared fusion behavior. Experiments report task results alongside source-copy controls and rendering sensitivity, tying each conclusion to its evaluation condition.

\begin{figure*}[!t]
\centering
\includegraphics[width=.88\linewidth]{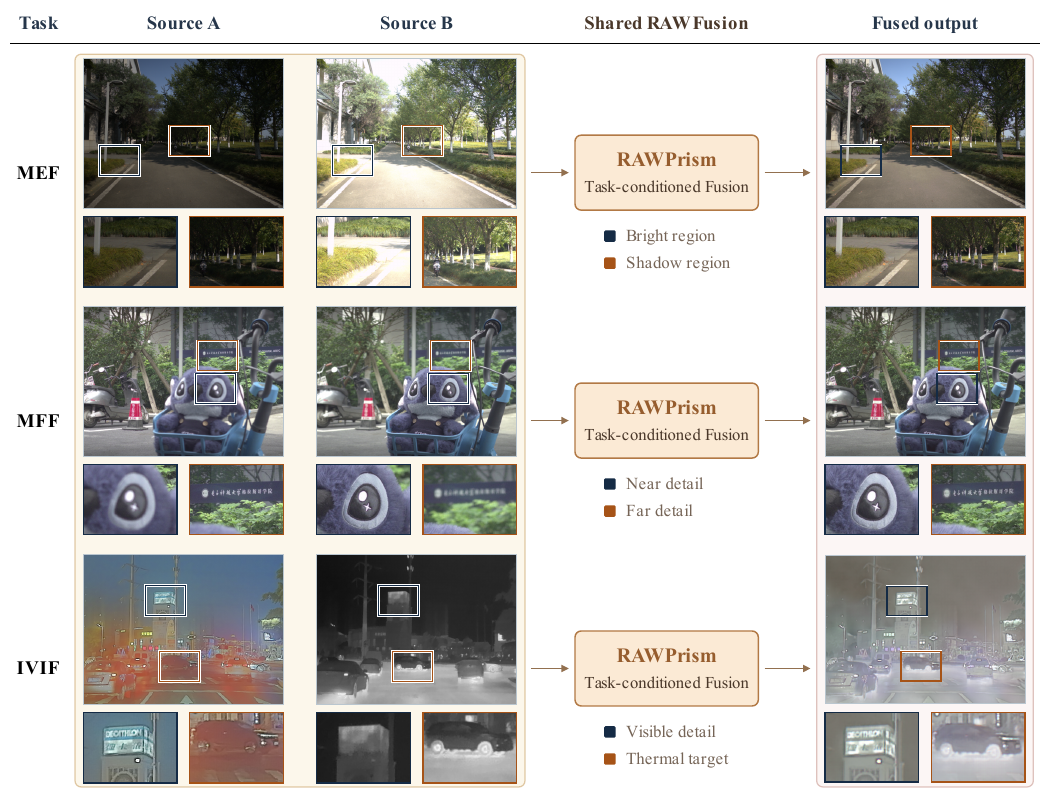}
\setcounter{figure}{0}
\caption{RAWPrism's three fusion settings. MEF pairs under/over exposures, MFF pairs near/far focus, and IVIF pairs reconstructed visible RAW with infrared. The examples illustrate complementary bright/dark detail, focus detail, and visible texture/thermal targets supplied to the shared model.}
\label{fig:1}
\end{figure*}

% Condensed related work; existing literature coverage retained.
\section{Related Work}
\label{sec:2}

\subsection{Image Fusion Datasets}
\label{sec:2.1}

{\looseness=1\emergencystretch=2em
SICE and MEFB support multi-exposure research \cite{ref2,ref3}; MFIFB, MFFW, and Real-MFF support multi-focus fusion \cite{ref4,ref5,ref6}. VIFB, LLVIP, M3FD, and FMB provide visible--infrared resources for fusion and related tasks \cite{ref1,ref7,ref8,ref9}. RAWPrism brings all three source configurations into the RAW domain through captured exposure pairs, captured and registered focus pairs, and FMB-derived reconstructed visible RAW. Section~\ref{sec:3} distinguishes their construction paths and provenance.\par}

\subsection{Fusion Methods and RAW-Domain Processing}
\label{sec:2.2}

Classical exposure fusion selects well-exposed regions using hand-crafted quality measures \cite{ref33}; UltraFusion handles large exposure differences through guided inpainting \cite{ref15}. Unified RGB methods share computation across tasks. U2Fusion uses unsupervised shared learning \cite{ref39}, SwinFusion models long-range cross-domain dependencies \cite{ref20}, and TC-MoA uses task-customized adapters \cite{ref21}. One Model for ALL exploits low-level task interaction \cite{ref22}. TITA combines task-invariant interaction with task-specific adaptation \cite{ref23}, while UniFusion combines robust representations with source-aware preservation \cite{ref24}.

RAW-domain processing includes learned ISPs and low-light restoration \cite{ref10,ref11}. Wang et al. address extreme low-light RAW enhancement and motion restoration with a domain-adaptive tokenizer and MeanFlow~\cite{ref41}. RawMEF reconstructs HDR from multiple exposures \cite{ref16}; RawMetaDiff uses dual exposures for low-light restoration \cite{ref17}. Burst methods support HDR and super-resolution reconstruction \cite{ref18,ref19}. Bayer-domain multi-focus fusion combines sources before demosaicing \cite{ref35}, while CFA-preserving preprocessing respects their sampling pattern \cite{ref36}. Unprocessing and inverse ISPs construct RAW-like inputs from RGB \cite{ref12,ref13,ref14}. ReRAW supplies our reconstructed visible inputs \cite{ref37}; CycleISP supplies their RGB counterparts \cite{ref38}. These works address individual tasks or ISP mappings. RAWPrism studies CFA-aware, task-conditioned fusion across MEF, MFF, and IVIF with one checkpoint.

\subsection{Fusion Evaluation}
\label{sec:2.3}

Fusion metrics can disagree with perceptual quality or downstream utility across focus, spectral, and exposure settings \cite{ref25,ref26,ref27}. Scores also depend on input provenance, rendering, spatial processing, and metric implementation. Our protocols declare these conditions and report complementary metrics with source-copy controls. Such controls retain one input without combining the pair, helping distinguish source similarity from successful fusion.

% Condensed dataset description; complete QC values and configuration checklists remain in Appendix B.
\section{The RAWPrism Dataset}
\label{sec:3}

\subsection{Design Goals and Organization}
\label{sec:3.1}

RAWPrism organizes exposure, focus, and spectral complementarity into three tracks (Table~\ref{tab:1}). MEF/MFF use captured RAW; IVIF uses reconstructed visible RAW paired with infrared. Paired observations share scene-level training, validation, and test partitions. Section~\ref{sec:method} defines their fusion, and Section~\ref{sec:setup} connects outputs to rendering and scoring.

% DOCX paragraphs: 60, 62, 63, 64, 65
\begin{table*}[!t]
\centering
\renewcommand{\thetable}{1}
\renewcommand{\theHtable}{source1}
\caption{The RAWPrism dataset: resources and scene-level splits. Resolutions reproduce the reported dataset dimensions. MEF retains 422 usable under/over pairs from 424 source records; two records are excluded during pairing and quality screening (Section~\ref{sec:3.2}).}
\label{tab:1}
\small
\setlength{\tabcolsep}{3pt}
\renewcommand{\arraystretch}{1.12}
\begin{tabularx}{\linewidth}{@{}p{.065\linewidth}Xp{.12\linewidth}p{.15\linewidth}p{.14\linewidth}@{}}
\toprule
Track & Source and construction & Usable scenes/pairs & Train / Val / Test & Reported resolution \\
\midrule
MEF & Our captured native RAW under/over exposure pairs & 422 & 338 / 42 / 42 & 4032 × 3024 \\
MFF & Our captured and registered native RAW focus pairs & 150 & 120 / 15 / 15 & 3990 × 2980 \\
IVIF & FMB-derived ReRAW reconstruction and CycleISP RGB rendering, each paired with registered thermal infrared & 1,500 & 1140 / 80 / 280 & 800 × 600 \\
\bottomrule
\end{tabularx}
\end{table*}

\subsection{Multi-Exposure Resource Construction}
\label{sec:3.2}

Each MEF scene provides under- and over-exposed RAW images with complementary bright- and dark-region detail. These two images supply the fusion inputs and source-derived training/validation objectives (Section~\ref{sec:training}); their renderings supply the scoring references (Section~\ref{sec:setup}). The track retains 422 usable pairs from 424 source records, split into 338 training, 42 validation, and 42 test scenes. Both inputs remain in the same partition. Appendix~\ref{sec:B} records the two exclusions; Appendix~\ref{sec:E.3} reports source-copy and averaging controls.

\subsection{Multi-Focus Resource Construction}
\label{sec:3.3}

Each MFF scene provides near- and far-focus RAW observations. Refocusing changes sharpness and image geometry, so registration establishes their correspondence before fusion. The track contains 150 registered pairs, split into 120 training, 15 validation, and 15 test pairs. Its spatially varying focus detail tests local source selection; Section~\ref{sec:setup} defines input and output rendering.

\subsection{Infrared--Visible Resource Construction}
\label{sec:3.4}

Starting from registered FMB visible--infrared pairs \cite{ref1}, ReRAW \cite{ref37} reconstructs visible RAW from the original RGB. CycleISP \cite{ref38} then renders that RAW as \path{rgbisp.jpg}. RAWPrism trains and tests on reconstructed RAW with infrared; RGB-domain models use the CycleISP rendering with the same infrared observation.

A separate OpenCV display renderer produces Protocol C's visible scoring reference and the fused RAW display. The reference uses reconstructed RAW before model-side visible gain (Section~\ref{sec:representation}). Thus CycleISP constructs RGB-model inputs, while OpenCV supplies evaluation images (Section~\ref{sec:setup}).

The track contains 1,500 scenes. We reserve 80 validation scenes from FMB's official 1,220-scene training set, leaving 1,140 for training and retaining its 280 test scenes. RAW, rendered RGB, and infrared share these partitions.

\subsection{Dataset Characterization and Quality Control}
\label{sec:3.5}

The 42-scene MEF test list matches the checkpoint's test partition and does not overlap its training or validation sets. Across the 422 retained MEF pairs, under-exposure saturation is at most 0.94\% (median 0.0\%); the exposure-time ratio spans 1.2x--356x (median 8.9x). All 150 MFF pairs retain full valid support after cropping, and complementary sharpness occurs in a median 36\% of analyzed tiles. IVIF files are complete across all 1,500 scenes, CFA packing is exactly invertible, and the CycleISP round trip reaches median PSNR/SSIM of 31.7 dB/0.96. Appendix~\ref{sec:B} reports the QC measurements and construction-record scope, including cross-partition near-duplicate candidates retained in the recorded splits.

\subsection{Resource Specification and Availability}
\label{sec:3.6}

We have prepared a reproducibility archive containing RAWPrism source code, inference and evaluation tools, configuration files, scene-partition records, and a final inference checkpoint. The archive's README and file manifest specify its contents and execution requirements. The full 24 GB image dataset, intermediate checkpoints, and ablation checkpoints are not included in this archive.

FMB imagery and derived visible RAW are identified separately with attribution. Appendix~\ref{sec:B} details the construction records and quality controls.

% Main-text compression, 2026-11-16; implementation details remain in Appendix D.
\section{RAWPrism: A Unified Three-Task Fusion Model}
\label{sec:method}

RAWPrism maps each input pair (Section~\ref{sec:3}) to CFA-aware latents, learns task-conditioned source weights, and reconstructs fused RAW through a shared backbone and task-specific head (Figure~\ref{fig:2}).

\subsection{Representation and Task-Conditioned Fusion}
\label{sec:representation}

\textbf{CFA-aware representation.} We pack the color filter array (CFA) samples as \(\mathbf X=[R,G_h,G_v,B]\in\mathbb R^{4\times H\times W}\), preserving both green sampling positions. A directional predictor combines horizontal and vertical green estimates with learned, gradient-aware weights. Its estimates \(\hat G_R,\hat G_B\) at red and blue sites define a common green reference:

\[
\begin{aligned}
\mathbf g&=[\hat G_R,G_h,G_v,\hat G_B], \\
\mathbf z&=\operatorname{cat}(\mathbf{H}_4\mathbf g,\ R-\hat G_R,\ B-\hat G_B) \\
&\in\mathbb R^{6\times H\times W},
\end{aligned}
\]

The fixed normalized Hadamard transform \(\mathbf H_4\) separates common green structure from inter-plane contrasts. Since \(\mathbf H_4^\top\mathbf H_4=\mathbf I\), these four channels and the two color residuals reconstruct the input to the decomposition (Appendix~\ref{sec:D.1}). The color residuals retain the differences between the original red/blue samples and their green estimates, so reconstructibility does not require perfect green prediction. This property concerns the input decomposition; IVIF subsequently removes the infrared color-difference channels.

\textbf{Task adapters.} For normalized under- and over-exposed MEF inputs \(U,O\), the adapter uses the scene-estimated over-to-under exposure ratio \(r\) to form \(U^\star=\sqrt r\,U\) and \(O^\star=O/\sqrt r\). It then restores canonical channel order before decomposition. MFF directly decomposes the normalized, registered pair.

IVIF packs normalized infrared samples by position within each \(2\times2\) cell. Each visible plane is scaled by the infrared-to-visible plane-mean ratio and clipped to \([0,1]\); \(\mathbf X_{\mathrm{vis}}\) denotes this gain-adjusted RAW throughout. Both inputs are decomposed, then the infrared latent's two color-difference channels are zeroed. Appendix~\ref{sec:D.5} specifies the preprocessing. Each adapter supplies six-channel latents \(\mathbf z_1,\mathbf z_2\).

\textbf{Source fusion.} The Reliability Field Gate (RFG) predicts spatial source preferences from both latents and a learned task condition \(\mathbf c_t\). Its convolutional network \(F_\theta\) uses GELU and feature-wise linear modulation (FiLM), followed by source-wise softmax:

\[
\begin{aligned}
(\boldsymbol\ell_1,\boldsymbol\ell_2) &=F_\theta\!\left(\operatorname{cat}(\mathbf z_1,\mathbf z_2),\mathbf c_t\right), \\
\mathbf w_i&=\frac{\exp(\boldsymbol\ell_i)}{\exp(\boldsymbol\ell_1)+\exp(\boldsymbol\ell_2)}, \quad i\in\{1,2\}.
\end{aligned}
\]

The maps sum to one and are broadcast across each source's six channels. Weighted concatenation retains both source streams:

\[
\begin{aligned}
\mathbf x_0&=\operatorname{cat}(\mathbf w_1\odot\mathbf z_1,\ \mathbf w_2\odot\mathbf z_2) \\
&\in\mathbb R^{12\times H\times W}.
\end{aligned}
\]

\subsection{Shared Backbone and Task-Specific Reconstruction}
\label{sec:reconstruction}

A shared three-level U-Net integrates \(\mathbf x_0\) using Restormer-style attention and gated feed-forward blocks~\cite{ref29}, with encoder--decoder skips. The task condition modulates features through adaptive LayerNorm. The backbone \(\mathcal B_\psi\) produces a six-channel latent, and head \(\mathcal H_t\) reconstructs four RAW planes:

\[
\begin{aligned}
\mathbf u&=\mathcal B_\psi(\mathbf x_0,\mathbf c_t), \\
\mathbf X_{\mathrm{head}}&=\mathcal H_t(\mathbf u).
\end{aligned}
\]

Each head applies a six-input, six-output \(1\times1\) residual projection to the backbone latent, adds the result to that latent, and performs inverse CFA reconstruction followed by a smooth range constraint. The residual projection is initialized to zero. Appendix~\ref{sec:D.1} specifies the synthesis and range functions. MEF/MFF use this output directly. MEF retains the shared exposure-reference scale without inverse scaling; the under-exposure DNG supplies its display template. Final IVIF inference combines the head with gain-adjusted visible RAW:

\[
\begin{aligned}
\mathbf X_{\mathrm{IVIF}} &=\operatorname{clip}_{[0,1]}\!\left((1-a)\mathbf X_{\mathrm{vis}}+a\mathbf X_{\mathrm{head}}\right), \\
a&=0.1.
\end{aligned}
\]

\begin{figure*}[!t]
\centering
\includegraphics[width=\linewidth,height=3.94in,keepaspectratio]{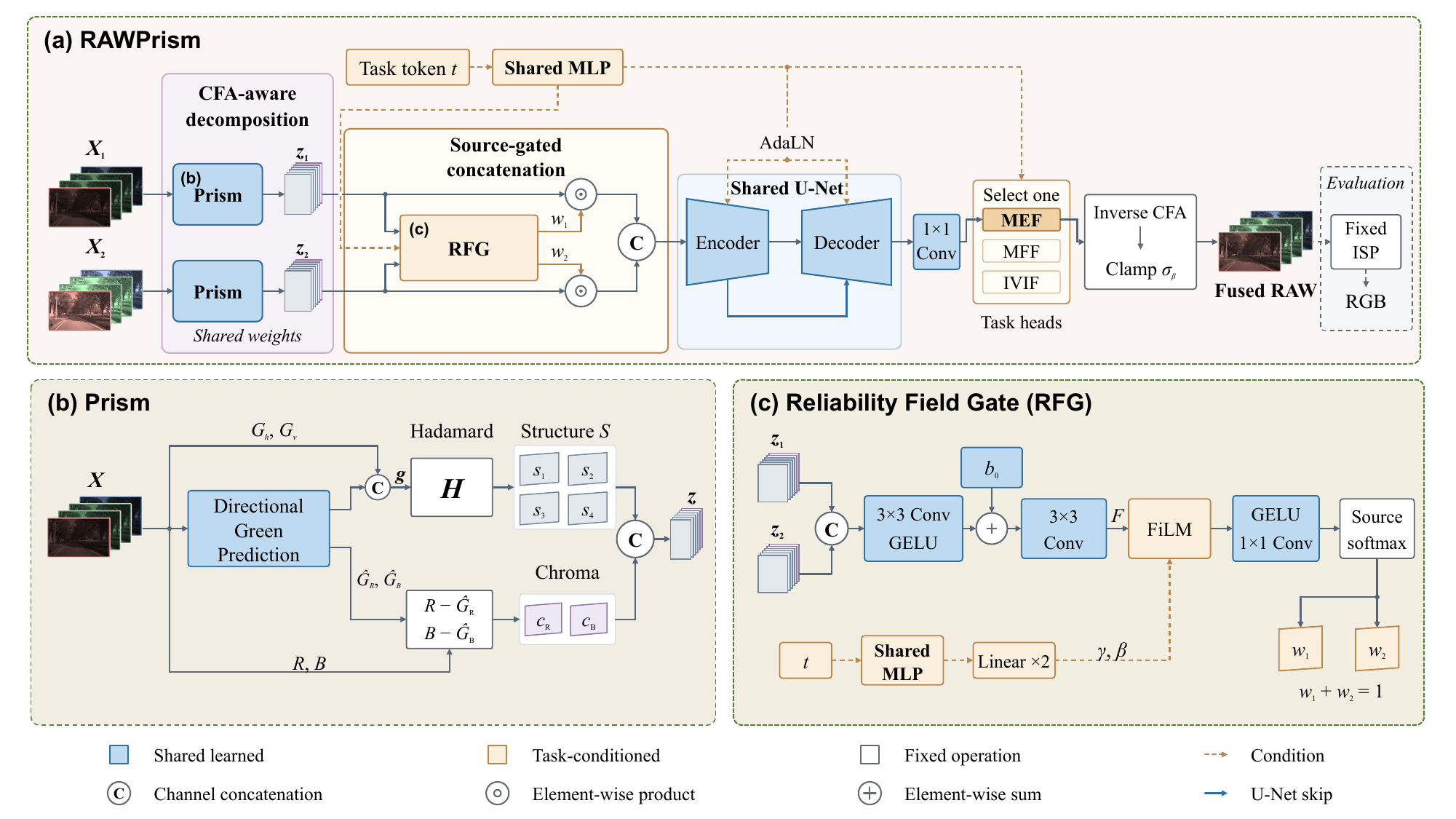}
\setcounter{figure}{1}
\caption{\textbf{RAWPrism architecture for unified RAW-domain MEF, MFF, and IVIF.}
\textbf{(a)} Task-specific input adapters provide two six-channel latents; omitted preprocessing includes MEF exposure alignment and IVIF infrared packing and visible-plane gain adjustment.
The thumbnails illustrate one MEF scene and an example output from a trained RAWPrism checkpoint; channel tints and a shared display mapping identify the packed CFA planes.
Spatial source weights are broadcast across each latent's six channels before weighted concatenation into the 12-channel shared U-Net input.
The selected head refines the six-channel backbone output as $\mathbf u_t=\mathbf u+\mathcal P_t(\mathbf u)$ using a task-specific $1\times1$ convolution, followed by inverse CFA synthesis and the smooth range constraint $\sigma_\beta$.
MEF/MFF use the head output directly; final IVIF inference additionally applies $\operatorname{clip}_{[0,1]}(0.9\mathbf X_{\mathrm{vis}}+0.1\mathbf X_{\mathrm{head}})$ (omitted), where $\mathbf X_{\mathrm{vis}}$ is gain-adjusted visible RAW.
The fixed display ISP serves RGB visualization/evaluation and differs from the differentiable training proxy.
\textbf{(b)} Directional Green Prediction combines horizontal and vertical estimates with learned, gradient-aware weights to obtain $\hat G_R,\hat G_B$ at red/blue sites.
Measured $G_h,G_v$ bypass prediction and are assembled with these estimates in the order $\mathbf g=[\hat G_R,G_h,G_v,\hat G_B]^\top$.
The fixed normalized Hadamard transform yields one common green component and three inter-plane contrasts; concatenating them with $R-\hat G_R$ and $B-\hat G_B$ forms the six-channel latent.
CFA analysis/synthesis is invertible before task-specific latent modifications; IVIF subsequently zeros the infrared color-difference channels.
\textbf{(c)} The shared MLP maps the task token to a condition that drives FiLM in the convolutional Reliability Field Gate (RFG) and AdaLN before backbone attention.
Source-wise softmax gives relative spatial preferences satisfying $\mathbf w_1+\mathbf w_2=1$.
In the reported configuration, hand-designed prior inputs are zero; the biased prior branch contributes the learned constant feature offset $b_0$.}
\label{fig:2}
\end{figure*}

\subsection{Training Objectives and Strategy}
\label{sec:training}

\textbf{RAW supervision.} All image targets derive from the paired sources; no fused ground truth is required. Joint training supervises \(\mathbf X_{\mathrm{head}}\); IVIF fine-tuning supervises the mixed \(\mathbf X_{\mathrm{IVIF}}\). The criteria compare reconstructed RAW intensity, structure, gradients, and chroma. Appendix~\ref{sec:D.2} specifies the individual penalties and stage-specific targets.

The task criteria use different forms of source complementarity. For MEF, the pixel target mixes the clipped exposure-aligned sources using relative-SNR and validity weights; structure and detail compare the prediction with weighted source SSIM and gradients. MFF computes focus weights from local gradient energy of each source's four-plane mean. Its sharpness term compares Gaussian high-pass magnitudes, alongside pixel, structure, gradient, and chroma penalties. IVIF operates on four-plane mean intensity for intensity, structure, and gradient terms, while preserving visible chroma. The documented adaptation objective selects the higher-activity source for intensity and structure, uses the maximum source-gradient magnitude, and includes visible and infrared intensity anchors. These source-derived targets specify the desired behavior independently of the learned gate.

\textbf{Rendering supervision.} A differentiable proxy \(\mathcal R_{\mathrm{train}}\) maps the supervised RAW to a tone-mapped three-channel image. It applies a RAW-to-three-channel mapping with learnable gains and biases and the tone curve \(h(x)=x/(1+x)\). The brightness target averages the two tone-mapped source intensities; the gradient target takes their maximum Sobel magnitude. Charbonnier penalties compare these targets with the rendered prediction. The documented rendering-loss weights are 0.3 for MEF/MFF and 1 for IVIF. MEF uses native normalized exposures before adapter alignment. During IVIF fine-tuning, the proxy renders the mixed prediction. This training proxy differs from CycleISP input construction and the display renderers used for scoring (Section~\ref{sec:setup}). Appendix~\ref{sec:D.2} specifies this auxiliary loss and the published joint configuration's semantic term.

\textbf{Training stages.} Joint training runs for 160 epochs, with 200 alternating updates per task per epoch and AdamW starting at \(10^{-4}\). The documented IVIF fine-tuning schedule specifies 60 epochs at \(5\times10^{-5}\), updating shared parameters and its adapter/head. Its documented residual coefficient is \(a=0.5\). We then merge only the fine-tuned IVIF adapter/head into the joint checkpoint, retaining the joint shared modules and MEF/MFF branches. Final inference uses \(a=0.1\). Configuration provenance and merge details are in Appendix~\ref{sec:D.4}.

\section{Benchmark Experiments and Analysis}
\label{sec:experiments}

One shared 3.58M-parameter RAWPrism checkpoint supports exposure, focus, and infrared--visible fusion. We evaluate its performance across these tasks, highlighting its strengths in MEF and IVIF and its applicability to MFF. Fixed-weight interventions and source controls further examine the representation and input responses underlying these results.

\subsection{Experimental Setup and Evaluation Protocol}
\label{sec:setup}

\textbf{Data and baselines.} We evaluate 42 MEF, 15 MFF, and 280 IVIF test scenes (Table~\ref{tab:1}). Baselines include SwinFusion~\cite{ref20}, UniFusion~\cite{ref24}, TC-MoA~\cite{ref21}, CDDFuse~\cite{ref30}, SeAFusion~\cite{ref31}, IFCNN~\cite{ref32}, Mertens~\cite{ref33}, and Laplacian fusion~\cite{ref34}; each table identifies the methods compared. MEF/MFF use target-trained RAWPrism and zero-shot learned baselines. IVIF methods train in their respective input domains with method-specific configurations. RGB models receive CycleISP images; RAWPrism receives reconstructed RAW. SwinFusion uses its official MEF ordering: A = underexposed, B = overexposed. The main method comparisons use one learned RAW model and one training seed.

\textbf{Rendering and scoring.} The evaluator receives \(F=\mathcal R_{\mathrm{eval}}(\widehat R)\) for RAW output or \(F=\widehat Y\) for RGB output. This display renderer differs from the training proxy in Section~\ref{sec:training}. IVIF renders the mixed \(\mathbf X_{\mathrm{IVIF}}\) through OpenCV. Protocol C scores against the OpenCV-rendered visible RAW before model-side gain and the infrared image. MEF retains the shared exposure scale and uses the under-exposure DNG template. MFF's primary evaluation uses the legacy R/B-swapped renderer.

\textbf{Metrics.} The metrics cover complementary properties: EN and SD describe output statistics; MI and SCD measure source-information association and cross-source difference correlations, respectively; \(Q_{AB/F}\) measures edge preservation; and VIF, CC, and SSIM assess source-referenced fidelity. The first five metrics use the output grid. VIF, CC, and SSIM average the two source comparisons on a maximum-side-1024 grid. Sources are globally resized only where permitted. MEF also uses three-level MEF-SSIM~\cite{ref28}. We report scene means and paired statistics on common scenes under the same condition. Appendix~\ref{sec:C} specifies the definitions and evaluation details.

\subsection{Multi-Exposure Fusion}
\label{sec:mef-results}

RAWPrism leads MI, SCD, and CC and ranks second in VIF (Table~\ref{tab:3}). Its paired MI gains are +1.788 [1.482, 2.058] over SwinFusion and +0.899 [0.601, 1.204] over Mertens, with wins on 40 and 35 of 42 scenes, respectively. These results highlight the shared model's strong exposure-fusion performance across several information and source-fidelity measures. SwinFusion and Mertens lead other criteria, illustrating the complementary properties captured by these measures. Appendix~\ref{sec:E.6} provides additional paired results.

% DOCX paragraphs: 196, 198, 199, 200, 201, 202, 203, 204
\begin{table*}[!t]
\centering
\renewcommand{\thetable}{2}
\renewcommand{\theHtable}{source3}
\caption{MEF results on 42 test scenes with a common evaluator, resize policy, and three-level MEF-SSIM. VIF, CC, and SSIM use the maximum-side-1024 grid. Source controls are discussed in Section~\ref{sec:controls}. Bold/underline: highest/second-highest values (ties included).}
\label{tab:3}
\small
\setlength{\tabcolsep}{3pt}
\renewcommand{\arraystretch}{1.06}
\begin{tabularx}{\linewidth}{@{}Xrrrrrrrrr@{}}
\toprule
Method & EN ↑ & SD ↑ & \(Q_{AB/F}\) ↑ & MI ↑ & SCD ↑ & VIF ↑ & CC ↑ & SSIM ↑ & MEF-SSIM ↑ \\
\midrule
SwinFusion, official order & 6.850 & 0.160 & \textbf{0.735} & 5.459 & 0.183 & 0.886 & 0.945 & \underline{0.636} & \underline{0.988} \\
UniFusion & 6.845 & 0.160 & 0.721 & 6.046 & 0.206 & 1.001 & \underline{0.953} & \textbf{0.639} & 0.980 \\
TC-MoA & \underline{6.898} & 0.138 & 0.651 & 3.448 & −0.268 & 0.733 & 0.851 & 0.595 & 0.922 \\
Mertens & \textbf{7.106} & \textbf{0.190} & \underline{0.729} & \underline{6.349} & \underline{0.217} & \textbf{1.039} & 0.937 & 0.632 & \textbf{0.989} \\
RAWPrism (ours) & 6.763 & \underline{0.161} & 0.443 & \textbf{7.248} & \textbf{0.380} & \underline{1.005} & \textbf{0.957} & 0.590 & 0.910 \\
\bottomrule
\end{tabularx}
\end{table*}

\subsection{Multi-Focus Fusion}
\label{sec:mff-results}

Under this primary MFF condition, RAWPrism obtains the highest mean MI of 5.758 and ranks third in edge preservation, with \(Q_{AB/F}=0.581\) (Table~\ref{tab:4a}). Its paired MI margin over SwinFusion is +0.102 [−0.17, 0.38]. These results support the shared framework's applicability to focus fusion. Appendix~\ref{sec:E.4} reports corrected-source sensitivity.

% DOCX paragraphs: 212, 214, 215, 216, 217, 218, 219, 220, 221, 222
\begin{table*}[!t]
\centering
\renewcommand{\thetable}{3}
\renewcommand{\theHtable}{source4a}
\caption{Primary MFF results on 15 scenes under the legacy R/B-swapped renderer. Sources are globally resized for FOV mismatch; VIF and SSIM use the maximum-side-1024 grid. Bold/underline: highest/second-highest values (ties included).}
\label{tab:4a}
\small
\setlength{\tabcolsep}{3pt}
\renewcommand{\arraystretch}{1.06}
\begin{tabularx}{\linewidth}{@{}Xrrrrrrr@{}}
\toprule
Method & EN ↑ & SD ↑ & \(Q_{AB/F}\) ↑ & MI ↑ & SCD ↑ & VIF ↑ & SSIM ↑ \\
\midrule
UniFusion & 6.699 & 0.121 & \underline{0.593} & 5.567 & 0.275 & 0.643 & \underline{0.910} \\
SwinFusion & 6.722 & 0.123 & 0.521 & \underline{5.656} & 0.472 & 0.638 & \textbf{0.921} \\
TC-MoA & 6.753 & 0.128 & 0.547 & 5.609 & 0.519 & \underline{0.684} & 0.896 \\
CDDFuse & 6.669 & \textbf{0.194} & 0.394 & 3.221 & \textbf{0.907} & 0.510 & 0.611 \\
IFCNN & \underline{6.760} & 0.128 & 0.558 & 5.613 & 0.631 & 0.633 & 0.900 \\
Laplacian & \textbf{6.802} & \underline{0.129} & \textbf{0.606} & 5.404 & \underline{0.681} & \textbf{0.698} & 0.889 \\
RAWPrism (ours) & 6.709 & 0.126 & 0.581 & \textbf{5.758} & 0.424 & 0.427 & 0.897 \\
\bottomrule
\end{tabularx}
\end{table*}

\subsection{Infrared--Visible Fusion}
\label{sec:ivif-results}

On 280 Protocol-C scenes, RAWPrism leads four of the eight reported metrics: EN, SD, MI, and VIF (Table~\ref{tab:5}). Its paired MI advantage over SwinFusion is +0.736 [0.647, 0.823]. The leading MI and VIF scores highlight its strength in both statistical information association and visual information fidelity. Together with the output-statistics results, they demonstrate strong performance across several cross-spectral evaluation dimensions within the same shared model. Edge, difference-correlation, and similarity measures yield different rankings. Appendix~\ref{sec:E.4} separately reports the standard-FMB scoring condition.

% DOCX paragraphs: 240, 242, 243, 244, 245, 246, 247, 248, 249, 250, 251
\begin{table*}[!t]
\centering
\renewcommand{\thetable}{4}
\renewcommand{\theHtable}{source5}
\caption{IVIF results on 280 Protocol-C scenes, scored against OpenCV-rendered visible RAW and infrared references. RAWPrism's fused RAW uses the same display pipeline. Bold/underline: highest/second-highest values (ties included).}
\label{tab:5}
\small
\setlength{\tabcolsep}{3pt}
\renewcommand{\arraystretch}{1.06}
\begin{tabularx}{\linewidth}{@{}Xrrrrrrrr@{}}
\toprule
Method & EN ↑ & SD ↑ & \(Q_{AB/F}\) ↑ & MI ↑ & SCD ↑ & VIF ↑ & CC ↑ & SSIM ↑ \\
\midrule
UniFusion & 7.240 & 0.202 & \textbf{0.716} & 4.352 & 1.596 & 0.419 & 0.586 & 0.718 \\
SwinFusion & 7.222 & 0.207 & 0.654 & \underline{4.548} & 1.672 & 0.382 & 0.601 & 0.727 \\
TC-MoA & \underline{7.249} & 0.200 & 0.703 & 3.884 & 1.563 & 0.409 & 0.581 & \textbf{0.735} \\
CDDFuse & 7.246 & \underline{0.213} & 0.645 & 4.512 & \textbf{1.717} & 0.394 & 0.604 & 0.732 \\
SeAFusion & 7.201 & 0.201 & 0.630 & 4.151 & \underline{1.679} & 0.367 & 0.605 & 0.718 \\
IFCNN & 7.069 & 0.171 & 0.648 & 3.543 & 1.547 & 0.327 & \underline{0.615} & \underline{0.733} \\
Laplacian & 7.189 & 0.173 & \underline{0.715} & 2.797 & 1.613 & \underline{0.432} & \textbf{0.636} & 0.730 \\
RAWPrism (ours) & \textbf{7.392} & \textbf{0.222} & 0.711 & \textbf{5.284} & 1.357 & \textbf{0.495} & 0.501 & 0.699 \\
\bottomrule
\end{tabularx}
\end{table*}

Figure~\ref{fig:3} provides qualitative examples for all three tasks under the display conditions stated in its caption, including Protocol S for IVIF.

\begin{figure*}[t]
\centering
\includegraphics[width=\linewidth]{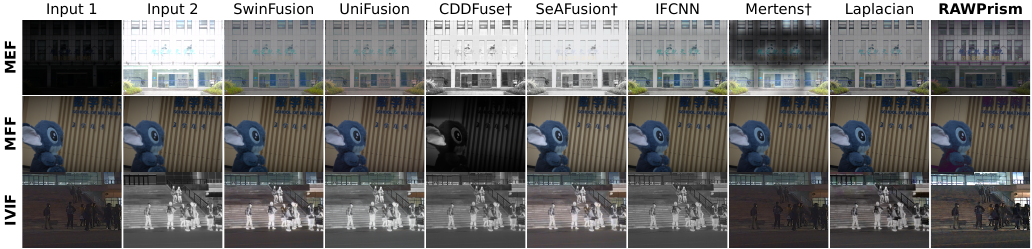}
\setcounter{figure}{2}
\caption{Qualitative comparisons of MEF, MFF, and IVIF (top to bottom) on scenes 0406, 0115, and 0060. The first two columns show the source pairs. IVIF follows Protocol S: RGB baselines receive original FMB visible RGB, while RAWPrism receives reconstructed RAW; both use the same infrared image. MFF inputs and the RAWPrism output use a shared metadata-based display ISP; Table~\ref{tab:4a} uses the legacy renderer. $\dagger$ marks cross-task applications: CDDFuse and SeAFusion on MEF/MFF, and Mertens on MFF/IVIF.}
\label{fig:3}
\end{figure*}

\subsection{Model Efficiency}
\label{sec:analysis}

RAWPrism combines three-task coverage with a compact 3.58M-parameter model, compared with 27.04M for UniFusion and 340.89M for TC-MoA. The evaluated SwinFusion variant uses 0.97M parameters. Appendix~\ref{sec:E.1} reports counting conventions and per-crop latency.

\subsection{Inference-Time Ablations}
\label{sec:ablations}

We probe the trained pipeline on 280 Protocol-C IVIF scenes using three representation interventions and three infrared-input perturbations (Table~\ref{tab:inference-ablations}). All settings retain the model weights and share the diagnostic replay reference. Representation controls change the indicated analysis or synthesis operation; infrared controls probe the complete input-processing and fusion path. Appendix~\ref{sec:E.ablations} specifies the replay scope and full infrared-control results.

% Existing reported fixed-weight study; all numeric cells retain their source values.
\begin{table*}[t]
\centering
\renewcommand{\thetable}{5}
\renewcommand{\theHtable}{inferenceablations}
\caption{Inference-time ablations on 280 Protocol-C IVIF scenes with fixed model weights. All rows share the diagnostic replay reference. Representation settings change the indicated operation; infrared settings probe the complete input-processing and fusion path. Full infrared metrics and execution scope are given in Appendix~\ref{sec:E.ablations}. Bold/underline: highest/second-highest values (ties included).}
\label{tab:inference-ablations}
\small
\setlength{\tabcolsep}{3pt}
\renewcommand{\arraystretch}{1.06}
\begin{tabularx}{\linewidth}{@{}Xrrrrrrr@{}}
\toprule
Inference-time setting & EN & SD & \(Q_{AB/F}\) & MI & SCD & VIF & SSIM \\
\midrule
Normal inference (reference) & 7.374 & 0.218 & \underline{0.716} & 5.187 & \textbf{1.389} & 0.489 & \textbf{0.704} \\
\midrule
Identity basis in place of Hadamard & \textbf{7.415} & \textbf{0.228} & 0.582 & 4.545 & \underline{1.088} & 0.364 & 0.623 \\
Fixed equal-weight green interpolation & 7.374 & 0.218 & \underline{0.716} & 5.189 & \textbf{1.389} & 0.489 & \textbf{0.704} \\
Four structure channels zeroed & 7.337 & \underline{0.221} & \textbf{0.724} & \textbf{6.949} & 0.769 & \textbf{0.519} & \textbf{0.704} \\
\midrule
Infrared zero & 7.356 & \underline{0.221} & \textbf{0.724} & 6.537 & 0.898 & \underline{0.514} & \underline{0.702} \\
Infrared mean & 7.342 & \underline{0.221} & \textbf{0.724} & \underline{6.645} & 0.789 & 0.512 & 0.701 \\
Infrared spatial shuffle & \underline{7.376} & 0.216 & 0.569 & 4.188 & 0.773 & 0.332 & 0.597 \\
\bottomrule
\end{tabularx}
\end{table*}

\textbf{Representation components.} The Hadamard coordinates are important to the trained computation: replacing both analysis and synthesis with identity lowers MI from 5.187 to 4.545, SCD from 1.389 to 1.088, \(Q_{AB/F}\) from 0.716 to 0.582, and VIF from 0.489 to 0.364. Equal-weight green interpolation leaves the aggregate scores nearly unchanged. Zeroing the four structure channels lowers SCD by 0.620 while increasing MI. These distinct responses show how the trained representation affects different aspects of the measured source relationships.

\textbf{Infrared content and arrangement.} Normal infrared input gives the highest SCD among these settings: 1.389, compared with 0.898, 0.789, and 0.773 for zero, mean, and spatially shuffled inputs. Shuffling also lowers \(Q_{AB/F}\) from 0.716 to 0.569 and VIF from 0.489 to 0.332, showing that the evaluated pipeline responds to spatially organized infrared content. Zero and mean inputs increase MI despite reducing SCD, illustrating why information scores and difference-correlation measures provide complementary evidence.

\subsection{Source and Protocol Controls}
\label{sec:controls}

Source controls provide additional evidence of RAWPrism's performance relative to simple copying and averaging (Appendix~\ref{sec:E.3}). Its MEF SCD of 0.380 exceeds rendered averaging at 0.221 and packed-RAW averaging at 0.264. For MFF, its SCD of 0.424 exceeds both source copies and their average. Under IVIF Protocol C, it obtains 1.357, compared with 0.723 for the visible copy and 0.677 for the infrared copy. These results show favorable difference-correlation performance against the stated controls.

The exact visible copy in IVIF obtains MI of 8.204 against RAWPrism's 5.284, despite its lower SCD. In MEF, copying the over-exposed source yields MI of 9.698 against 7.248 (Appendix~\ref{sec:E.3}). These scores show why we interpret MI alongside difference correlation and edge measures.

Under standard FMB Protocol S, RAWPrism's MI falls from 5.284 to 3.492 and VIF from 0.495 to 0.312. Relative to Protocol C, both the visible scoring reference and RGB baseline inputs change (Appendix~\ref{sec:E.4}).

\section{Discussion and Limitations}
\label{sec:discussion}

MEF and MFF use captured RAW, while IVIF uses inverse-ISP reconstructions. Rankings reflect the reported training conditions, including zero-shot learned baselines for MEF/MFF and method-specific training for IVIF. The test sets include 15 MFF scenes. The corrected-source MFF results show that method rankings can change on those same scenes under a different rendering protocol (Appendix~\ref{sec:E.4}). Paired intervals capture scene variation for checkpoints from a single training seed.

Fixed-weight interventions characterize the inference computation. Comparing captured RAW with its rendered RGB counterpart on the same scenes, under matched training and scoring conditions, would isolate the contribution of input representation. Training the same backbone separately for each task would quantify the effect of sharing parameters.

For IVIF, the visible-plane gain depends on the infrared input (Section~\ref{sec:representation}). The replay records leave the gain treatment unspecified, so the infrared perturbations characterize the full preprocessing and fusion pipeline. Holding gain fixed during these perturbations and scoring the gain-adjusted \mbox{visible input} alongside the fused output would help isolate the contribution of the fusion branch.

The audit flagged cross-split near-duplicate candidates from the same capture series (Appendix~\ref{sec:B}). Grouping these series before partitioning would provide a stricter test of generalization to independent captures. Additional camera conditions and training seeds would help quantify variation across acquisition and training settings.

\section{Conclusion}
\label{sec:conclusion}

RAWPrism-Benchmark connects captured exposure/focus pairs and reconstructed visible--thermal pairs to scene-level evaluation. RAWPrism combines CFA-aware encoding, task-conditioned source selection, and a shared backbone in one 3.58M-parameter checkpoint. Together, they support unified RAW-input fusion while exposing task-dependent metric trade-offs and sensitivity to rendering and scoring conditions.

The reproducibility materials and their scope are described in Section~\ref{sec:3.6}.

% DOCX paragraphs: 277
\section*{LLM Usage Statement}

% DOCX paragraphs: 279
OpenAI tools (ChatGPT / GPT) were used for coding assistance, experiment scripting, manuscript editing, LaTeX formatting, and reference checking. The authors are responsible for verifying the experiments, references, and claims and retain responsibility for the content of the paper.
\par

% Official CVPR bibliography style; metadata is maintained in references.bib.
{\small
% Include the supplementary-only references to keep one shared numbering.
\nocite{*}
\bibliographystyle{ieeenat_fullname}
\bibliography{references}
}

\clearpage
% DOCX paragraphs: 358
\section*{Supplementary Material}

\appendix
% DOCX paragraphs: 360
\section{Resource Specification and Scene Partitions}
\label{sec:A}

% DOCX paragraphs: 362
The training/validation/test splits contain 338/42/42 MEF scenes, 120/15/15 MFF pairs, and 1140/80/280 IVIF scenes. The MEF test set matches the joint checkpoint's 42-scene test split. For IVIF, we kept FMB's official test set and drew the validation scenes from its training set. We retained 422 under/over pairs from 424 MEF source records. Scene \texttt{scene\_0009} lacked extracted top-level DNG artifacts, and \texttt{scene\_0120} lacked an under/over pair (Appendix~\ref{sec:B}).

Scene-partition files and verification records are included in the reproducibility archive described in Section~\ref{sec:3.6}. We plan to release the RAWPrism dataset together with our additional Sony $\alpha$7~III multi-exposure and multi-focus captures (Appendix~\ref{sec:G}) and publish the Baidu Netdisk download link.

% DOCX paragraphs: 366
\section{Construction and Data Quality}
\label{sec:B}

% DOCX paragraphs: 368
We captured exposure pairs for MEF and focus pairs for MFF, then registered the focus pairs. For IVIF, ReRAW reconstructs visible RAW for RAWPrism, and CycleISP renders the corresponding RGB for RGB-domain methods. The two representations share scene partitions for training and testing. For each scene, both pair with the same infrared image.

\begin{table*}[t]
\noindent Excluded MEF source records
\par\smallskip
\centering
\small
\setlength{\tabcolsep}{4pt}
\renewcommand{\arraystretch}{1.12}
\begin{tabularx}{\textwidth}{@{}p{.12\textwidth}p{.13\textwidth}p{.34\textwidth}X@{}}
\toprule
scene\_id & batch & exclusion reason & action \\
\midrule
scene\_0009 & DatasetMEF & no extracted top-level DNG artifacts or meta.json & excluded \\
scene\_0120 & DatasetMEF & no under/over pair & excluded, metadata record retained \\
\bottomrule
\end{tabularx}
\end{table*}

% DOCX paragraphs: 370

% Full construction/QC details moved from the main paper during eight-page editing.
\subsection{Detailed Quality Controls and Construction Records}
\label{sec:B.details}

We checked DNG decoding for all 422 retained MEF scenes. Each scene contains under- and over-exposed RAW observations with exposure metadata. After black-level correction, at most 0.94\% of pixels saturate in under-exposed observations (median 0.0\%). Over-exposed observations have a median shadow-content fraction of 4.0\%. The exposure-time ratio ranges from \(1.2\times\) to \(356\times\) (median \(8.9\times\)).

All 150 registered MFF pairs retain full valid support after cropping. The dataset records include per-pair registration residuals. All required files are present for the 1,500 FMB-derived IVIF scenes, and the CFA pack/unpack transform is exactly invertible.

We screened all three tracks for cross-split near-duplicates using 64-bit pHash and a Hamming distance of at most 10. The dataset metadata lists the candidates, most of which pair adjacent captures from the same series (Section~\ref{sec:3.5}). We kept the evaluation splits after the audit, including FMB's official test set for IVIF.

\textbf{Construction-record scope.}

The benchmark records are incomplete for camera and lens settings, capture procedures, MFF registration and sample selection, and the ReRAW/CycleISP checkpoints, conversion settings, and JPEG export. Appendix~\ref{sec:D.5} specifies the model's input normalization, packing, and exposure scaling.

\textbf{Supplementary-material scope.}

The archive contains source code, inference and evaluation tools, configurations, scene-partition files, the final inference checkpoint, and verification records. Its README and file manifest list the contents and execution requirements. Appendix~\ref{sec:F} shows source images and fusion outputs, including visible and infrared examples from FMB~\cite{ref1}, and Appendix~\ref{sec:G} shows Sony captures. The full image datasets, individual FMB and reconstructed RAW files, complete per-scene records, the v10 joint checkpoint, the IVIF-branch delta, and ablation checkpoints are outside the archive.

% DOCX paragraphs: 372
\section{Evaluation Implementation}
\label{sec:C}

\subsection{Input Representations and Rendering}
\label{sec:C.1}

For scene \(i\), let \(V_i\) and \(T_i\) denote the original FMB visible RGB and infrared images. ReRAW~\cite{ref37} reconstructs visible RAW \(R_i\), and CycleISP~\cite{ref38} renders it as RGB \(C_i\) (Section~\ref{sec:3.4}). Pairing each visible representation with \(T_i\) gives the IVIF inputs before method-specific normalization and packing.

\[
\begin{aligned}
V_i &\xrightarrow{\mathrm{ReRAW}} R_i
    \xrightarrow{\mathrm{CycleISP}} C_i,\\
X_i^{\mathrm{RAW}} &=(R_i,T_i),\qquad
X_i^{\mathrm{RGB}}=(C_i,T_i).
\end{aligned}
\]

RAWPrism trains and tests on \(X_i^{\mathrm{RAW}}\). RGB-domain learned methods use \(X_i^{\mathrm{RGB}}\) on the same scene partitions, with \(C_i\) stored as \texttt{rgbisp.jpg}. RAWPrism uses a differentiable rendering proxy for training supervision (Section~\ref{sec:training} and Appendix~\ref{sec:D.2}). Within each IVIF protocol, RAWPrism and the RGB methods share the visible and infrared scoring references. The evaluator scores the RGB methods on their RGB outputs.

For IVIF, the inference implementation saves the mixed four-plane output \(\mathbf X_{\mathrm{IVIF}}\) defined in Section~\ref{sec:reconstruction} and renders the evaluation image through \path{render_ivif_display} in \path{tools/infer_remote.py}. This function calls \path{render_raw4} in \path{src/finalfusion/isp_rawpy.py}, which reconstructs the Bayer mosaic, quantizes it to 16-bit values, and demosaics it with OpenCV. The IVIF configuration uses neutral white balance and an identity color matrix, followed by sRGB transfer. The display function applies gray-world white balance and a 0.5/99.5 percentile stretch based on the mean RGB intensity, then saves the evaluation image as an 8-bit PNG. The recorded Protocol C command selects the \path{legacy} renderer, which retains the historical R/B channel mapping.

% DOCX paragraphs: 388
Protocol C uses the references \(A_i^{\mathrm C}=\mathcal R_{\mathrm{eval}}(R_i)\) and \(B_i^{\mathrm C}=T_i\). The evaluation renderer \(\mathcal R_{\mathrm{eval}}\) uses OpenCV to render the normalized reconstructed visible RAW before model-side gain adjustment. The fused prediction uses the gain-adjusted input described in Appendix~\ref{sec:D.5}. The visible reference is documented as \path{vis_rgb_rt.png} and stored at \path{data_ivif_rt/ivif/test/<scene>/vis_rgb.png} for evaluation.

MEF exports its prediction at the shared exposure scale using the under-exposure DNG template (Appendix~\ref{sec:D.5}). MFF uses the legacy renderer for its primary comparison, paired analysis, and source controls. The MFF sensitivity analysis uses corrected-source rendering (Table~\ref{tab:C1}).

% Expanded source Table~\ref{tab:C1}; existing QC and current result references retained.
\begin{table*}[!t]
\centering
\renewcommand{\thetable}{C1}
\renewcommand{\theHtable}{sourceC1}
\caption{Input and evaluation conditions. IVIF result provenance is discussed with Table~\ref{tab:5}.}
\label{tab:C1}
\small
\setlength{\tabcolsep}{3pt}
\renewcommand{\arraystretch}{1.12}
\begin{tabularx}{\linewidth}{@{}p{.13\linewidth}p{.25\linewidth}p{.27\linewidth}X@{}}
\toprule
Track / condition & RAWPrism input & RGB-baseline input specified by the protocol & Scoring and spatial processing \\
\midrule
MEF & Captured under/over RAW pair & Fixed renderings of the under/over RAW inputs & Rendered under/over references; source resizing permitted for FOV mismatch; three-level MEF-SSIM \\
MFF, primary (legacy) & Captured, registered RAW focus pair & Source renderings under the legacy R/B-swapped pipeline & Legacy rendered sources; global source resizing when dimensions differ \\
MFF, corrected-source sensitivity & Captured RAW focus pair & Corrected source renderings & Corrected channel order and scene white balance. Table~\ref{tab:E4} reports results. Run linkage and baseline input identities are only partially recorded (Appendix~\ref{sec:E.4}). \\
IVIF, ReRAW--CycleISP inputs & ReRAW-reconstructed visible RAW and corresponding thermal infrared input & CycleISP-rendered \path{rgbisp.jpg} and the same thermal input, for training and testing & Shared scene partitions; OpenCV display ISP for fused RAW; scoring references specified by the evaluation condition \\
\bottomrule
\end{tabularx}
\end{table*}

\subsection{Spatial Support and Metric Definitions}
\label{sec:C.2}

Let \(A_i,B_i\) be the scoring references for the evaluation condition and \(F_i^{(k)}\) the rendered output of method \(k\) for scene \(i\). When the protocol permits resizing, we resize each source globally to match the fused output dimensions. The source and output dimensions match for all 42 MEF test scenes. We use global resizing for MFF when dimensions differ.

We report source-averaged PSNR (\(\mathrm{PSNR}_{\mathrm{src}}\)) for fusion outputs in the supplementary analyses. We compute VIF, CC, SSIM, and PSNR on a grid with maximum side length 1024. For output dimensions \(H\times W\), the scale is \(s=\min(1,1024/\max(H,W))\), with dimensions rounded to integers. Images whose longest side is at most 1024 retain their dimensions. Write \(\widetilde A_i^{(k)},\widetilde B_i^{(k)},\widetilde F_i^{(k)}\) for the resulting images after this per-method spatial processing and the metric's grayscale conversion. For \(m\in\{\mathrm{VIF},\mathrm{CC},\mathrm{SSIM},\mathrm{PSNR}\}\), the reported per-scene value is

\[
q_{i,m}^{(k)}=\frac12\left[
 m\!\left(\widetilde A_i^{(k)},\widetilde F_i^{(k)}\right)
+m\!\left(\widetilde B_i^{(k)},\widetilde F_i^{(k)}\right)
\right].
\]

The source is the first argument and the fused image the second. EN, SD, \(Q_{AB/F}\), MI, and SCD use the evaluation output grid. EN and SD measure the fused image. MI sums its mutual information with the two sources, and \(Q_{AB/F}\) and SCD use their respective three-image definitions. MEF also uses the same three-level MEF-SSIM implementation across methods~\cite{ref28}.

An exact source copy yields infinite PSNR against that input. A high source-referenced score can come from matching one input alone. Section~\ref{sec:controls} and Appendix~\ref{sec:E.3} report the source-copy controls.

\subsection{Scene Aggregation and Paired Comparisons}
\label{sec:C.3}

We aggregate scores separately for each task and rendering condition. Let \(\mathcal I_k\) be method \(k\)'s evaluated scene set and \(q_{i,m}^{(k)}\) its value for metric \(m\), computed using the corresponding definition in Section~\ref{sec:C.2}. The scene mean and paired mean difference from comparison method \(b\) are

\[
\begin{aligned}
\overline q_m^{(k)}
&=\frac{1}{|\mathcal I_k|}
  \sum_{i\in\mathcal I_k}q_{i,m}^{(k)},\\
\Delta_m^{(k,b)}
&=\frac{1}{|\mathcal I_k\cap\mathcal I_b|}
  \sum_{i\in\mathcal I_k\cap\mathcal I_b}
  \left(q_{i,m}^{(k)}-q_{i,m}^{(b)}\right).
\end{aligned}
\]

We report the number of evaluated scenes for each comparison. Paired statistics use the common scenes under the same evaluation condition. We bootstrap the paired differences to obtain intervals for the mean difference and count scenes with positive differences.

\subsection{Metric Implementation Conventions}
\label{sec:C.4}

The reported \(Q_{AB/F}\) implementation uses relative source--fused gradient strength and a \(\pi\)-periodic orientation term.

SCD uses mean-centered correlation of the cross-source differences:

\[
\begin{aligned}
\operatorname{SCD}(F,A,B)={}&\operatorname{corr}(F-B,A) \\
&+\operatorname{corr}(F-A,B).
\end{aligned}
\]

VIFp accumulates its numerator and denominator across scales before forming their ratio. We use these definitions for method comparisons and source controls and report scores separately for each rendering and input-reference condition.

For some baselines, the archived records only partially link the inputs and outputs to exact evaluator and renderer versions. Appendix~\ref{sec:E.4} describes the archived corrected-source MFF and standard-FMB IVIF conditions.

% Implementation-to-objective connections checked against available code and configurations.
\section{RAWPrism Implementation}
\label{sec:D}

\subsection{CFA Representation, Backbone, and Reconstruction}
\label{sec:D.1}

For the ordering \(\mathbf g=[\hat G_R,G_h,G_v,\hat G_B]^\top\), the normalized Hadamard matrix is

\[
\mathbf H_4=\frac12
\begin{bmatrix}
1&1&1&1\\
1&-1&1&-1\\
1&1&-1&-1\\
1&-1&-1&1
\end{bmatrix},
\qquad \mathbf H_4^\top\mathbf H_4=\mathbf I.
\]

Its first coefficient represents a common component of the four green-related values; the other three represent signed differences across the CFA positions. The transform changes the basis without reducing its dimension. If \(\mathbf z_{1:4}\) denotes the four structure channels and \(z_5,z_6\) the color differences, input reconstruction is

\[
\begin{aligned}
\mathbf g&=\mathbf H_4^\top\mathbf z_{1:4},\\
R&=g_1+z_5,\qquad G_h=g_2,\\
G_v&=g_3,\qquad B=g_4+z_6.
\end{aligned}
\]

Thus the analysis \(\mathcal A\) and synthesis \(\mathcal S\) satisfy \(\mathcal S(\mathcal A(\mathbf X))=\mathbf X\) before any task-specific modification of the encoded latents. Here \(\mathbf X\) is the tensor supplied to CFA analysis after any input scaling or channel reordering. The red and blue values remain recoverable through their residuals even when the green estimates are imperfect.

For the infrared input, the four packed planes contain the top-left, top-right, bottom-left, and bottom-right grayscale samples from each \(2\times2\) cell. The \(R,G_h,G_v,B\) labels identify these spatial positions when the same decomposition is applied to infrared. Zeroing the two infrared color-difference channels discards local information at the R/B positions.

The source softmax assigns relative RFG weights without noise-probability calibration. The weights sum to one even when both inputs are unreliable. In the reported configuration, the prior branch takes zero inputs in place of the hand-designed reliability priors. Its biased convolutions can still contribute a learned constant offset.

The backbone uses three resolutions with feature widths of 64, 128, and 256. Its two encoder stages, bottleneck, and two decoder stages each contain two transformer blocks with eight attention heads per block. Within each head, channel-wise attention forms a \((C/h)\times(C/h)\) matrix. A learned task token produces adaptive LayerNorm scales and shifts.

At the task head, the shared backbone output \(\mathbf u\in\mathbb R^{6\times H\times W}\) is refined as

\[
\mathbf u_t=\mathbf u+\mathcal P_t(\mathbf u),
\]

where \(\mathcal P_t\) is a task-specific \(1\times1\) convolution with six input and six output channels and zero-initialized weights and bias. CFA synthesis reconstructs the RAW planes from \(\mathbf u_t\) using the inverse Hadamard transform and the two color-difference channels. The MEF head restores its external channel order.

After inversion to RAW planes, the smooth range constraint is

\[
\begin{aligned}
\sigma_\beta(x)={}&\operatorname{softplus}(x,\beta) \\
&-\operatorname{softplus}(x-1,\beta).
\end{aligned}
\]

\subsection{Supervised Outputs and Source-Derived Objectives}
\label{sec:D.2}

The task criterion and differentiable rendering proxy use the same four-plane prediction \(P\). Joint training uses \(P=\mathbf X_{\mathrm{head}}\). When the IVIF visible residual is enabled, both receive only the mixed prediction \(P=\mathbf X_{\mathrm{IVIF}}\). The paired sources supply the targets and their reliability, focus, or activity weights. The source gate uses learned weights.

\textbf{MEF.} The pixel target is the reliability-weighted mixture of \(\operatorname{clip}_{[0,1]}(U^\star)\) and \(\operatorname{clip}_{[0,1]}(O^\star)\). Structure is a weighted sum of SSIM comparisons with these two references, and detail compares the prediction's per-plane gradients with the weighted source gradients. Chroma compares red/blue differences from mean green with those of the pixel target. A range penalty additionally acts on the reconstructed RAW before the smooth range constraint. Reliability uses the native under/over observations and the fitted four-channel noise model; the supervision references use the common exposure scale defined in Section~\ref{sec:representation}.

\textbf{MFF.} The sources are the normalized, registered RAW pair. Local gradient energy of each source's four-plane mean determines its focus weight. Pixel and SSIM penalties compare \(P\) separately with both sources and weight the two penalties spatially. Gradient, chroma, and sharpness penalties compare the corresponding prediction quantities with weighted source quantities. Sharpness is the magnitude of the difference between the mean-intensity field and its Gaussian-blurred version, using a \(5\times5\) kernel with \(\sigma=1\).

\textbf{IVIF.} Intensity, SSIM, and gradient terms operate on the four-plane means of \(P\), gain-adjusted \(\mathbf X_{\mathrm{vis}}\), and packed infrared. Chroma compares \(P\) with \(\mathbf X_{\mathrm{vis}}\) only. The joint-training configuration uses soft weights from local squared-gradient activity for intensity, structure, and gradient targets. The documented IVIF fine-tuning configuration instead selects the higher-activity source for the intensity target and structure comparison, uses the maximum source-gradient magnitude as the gradient target, and adds visible/infrared intensity anchors with weights 1 and 0.5. Both configurations penalize excessive predicted gradients in regions where both sources are flat. Their intensity, structure, gradient, chroma, and artifact coefficients are 2, 1, 1, 1, and 0.1, respectively.

\textbf{Differentiable rendering.} The training proxy \(\mathcal R_{\mathrm{train}}\) maps packed RAW to three channels on the packed spatial grid, applies learnable channel gain and bias with clipping, and uses the tone curve \(h(x)=x/(1+x)\). The full-resolution display ISP is described in Section~\ref{sec:C.1}. If \(S_1,S_2\) are the source tensors passed to the loss and \(\mu_c\) denotes a channel mean, the rendering targets and loss are

\[
\begin{aligned}
 t_i &= h(\mu_c(S_i)),\qquad y=\mu_c(\mathcal R_{\mathrm{train}}(P)),\\
 \mathcal L_{\mathrm{render}} &= \big\langle\rho\big(y-\tfrac12(t_1+t_2)\big)\big\rangle\\
 &\quad+\big\langle\rho\big(G(y)-\max(G(t_1),G(t_2))\big)\big\rangle,
\end{aligned}
\]

where \(G\) is Sobel gradient magnitude, \(\rho\) is the Charbonnier penalty, and brackets denote spatial averaging. For MEF, the rendering loss uses the normalized native under/over tensors before the adapter's exposure scaling, and the RAW task loss uses the aligned references described above. MFF uses the registered pair, and IVIF uses gain-adjusted visible RAW and packed infrared. The rendering loss is added to the RAW task objective with weight 0.3 for MEF/MFF and 1 for IVIF in the documented configurations.

\textbf{Semantic regularization.} The source joint-training configuration assigns weight 0.5 to an IVIF feature loss with a frozen DINOv3~\cite{ref40} ViT-S/16. The loss compares features of the differentiably rendered prediction with those of the rendered gain-adjusted visible input and replicated infrared intensity. The staged joint-training and IVIF fine-tuning configurations set this weight to zero.

\subsection{Training Stages and Checkpoint Construction}
\label{sec:D.4}

The joint-training configuration specifies 160 epochs with 200 alternating updates per task per epoch (Section~\ref{sec:training}). It uses AdamW with initial learning rate \(10^{-4}\), gradient clipping at 5.0, ReduceLROnPlateau, and bfloat16 mixed precision on one NVIDIA RTX 4090 D with 24GB memory. The reported seed is 20260816.

The IVIF fine-tuning configuration specifies 60 epochs at a learning rate of \(5\times10^{-5}\), a visible-residual coefficient of \(a=0.5\), and the source-derived objectives in Section~\ref{sec:D.2}. Fine-tuning updates the shared modules and IVIF adapter/head. Checkpoint assembly then copies only the adapted IVIF adapter and head into the joint model; all shared parameters and MEF/MFF branches retain their joint-training values. Final IVIF inference uses \(a=0.1\) as a fixed deployment setting.

The archived tensor-level verification confirms the branch merge, residual coefficient, and strict loading for the checkpoint identified in that record. The supplementary archive separately records loading and forward-pass checks for the included inference checkpoint. The configuration records specify the implemented fine-tuning procedure. A complete training log that ties every setting to the adapted IVIF checkpoint is unavailable. Validation optimality for the fixed inference coefficient remains unverified. Section~\ref{sec:3.6} lists the available assets.

\subsection{Input Normalization and Exposure Scales}
\label{sec:D.5}

\textbf{MEF exposure reference.} Native RAW values are black-level corrected and normalized by the sensor range. The exposure ratio \(r\) is estimated from the pair's overlapping valid brightness range, using a trimmed mean in log space. CFA decomposition receives the unclipped, scaled inputs \(U^\star=\sqrt r\,U\) and \(O^\star=O/\sqrt r\). The RAW loss clips its source references to \([0,1]\). Export unpacks the predicted RAW and restores digital values at the common exposure scale using the black and white levels. The under-exposure DNG supplies the display-rendering template.

\textbf{IVIF visible gain.} Let \(V_c\) and \(I_c\) be normalized visible and infrared planes before model-side gain adjustment. With full-scene spatial means \(\overline V_c,\overline I_c\), preprocessing computes

\[
g_c=\frac{\overline I_c}{\overline V_c+10^{-6}},\qquad
\mathbf X_{\mathrm{vis},c}=\operatorname{clip}_{[0,1]}(g_cV_c).
\]

Training computes these gains before crop sampling; inference computes them before tiling. The resulting \(\mathbf X_{\mathrm{vis}}\) enters the adapter, the direct visible residual, and the source-derived losses. Protocol C instead renders \(V\), before this gain adjustment, as its visible scoring reference (Appendix~\ref{sec:C}). The four infrared planes retain the four spatial sub-lattices of the grayscale image.

% Appendix~\ref{sec:E} synchronized with the author-maintained expanded manuscript.
\section{Additional Analysis and Runtime}
\label{sec:E}

\subsection{Model Size and Runtime}
\label{sec:E.1}

We count shared parameters once. RAWPrism contains 3.58M parameters, compared with 27.04M for UniFusion, 340.89M for TC-MoA, and 0.97M for the evaluated SwinFusion variant. UniFusion has 5.44M trainable parameters. TITA reports 1.39M parameters with luminance input~\cite{ref23}.

We measure crop latency on an otherwise idle RTX 4090 D with PyTorch 2.8.0+cu128, fp32, and batch size 1. CUDA events time 200 iterations after 50 warm-up iterations. Table~\ref{tab:E1} lists the square-crop side lengths in packed and sensor pixels. The timing covers crop inference only. Full-resolution tiling and end-to-end image processing fall outside the timing window.

% DOCX paragraphs: 423, 425, 426, 427, 428, 429
\begin{table}[!t]
\centering
\renewcommand{\thetable}{E1}
\renewcommand{\theHtable}{sourceE1}
\caption{Latency in milliseconds for square crops, with side lengths in pixels.}
\label{tab:E1}
\small
\setlength{\tabcolsep}{3pt}
\renewcommand{\arraystretch}{1.12}
\begin{tabularx}{\linewidth}{@{}Xrrrr@{}}
\toprule
Task & Packed side & Sensor side & Median latency & p95 latency \\
\midrule
MEF & 128 & 256 & 13.9 & 14.8 \\
MFF & 256 & 512 & 37.5 & 38.7 \\
IVIF & 128 & 256 & 11.6 & 13.9 \\
\bottomrule
\end{tabularx}
\end{table}

\subsection{Inference-Time Ablation Details}
\label{sec:E.ablations}

\textbf{Experimental scope.} We compare all interventions within a separate fixed-weight replay on 280 Protocol-C scenes. The replay reference MI/SCD is 5.187/1.389. Table~\ref{tab:5} reports 5.284/1.357, and the records do not establish a match between the two runs. The replay record does not say whether the infrared interventions held the visible gain fixed or recomputed it. The infrared interventions probe the pipeline from preprocessing through fusion.

\textbf{Representation interventions.} We replace both Hadamard analysis and synthesis with identity, substitute equal-weight green interpolation, or zero the four structure channels (Table~\ref{tab:inference-ablations}). Identity lowers MI/SCD/\(Q_{AB/F}\)/VIF to 4.545/1.088/0.582/0.364. Equal-weight interpolation changes MI by +0.002. Structure-channel zeroing lowers SCD to 0.769 while increasing MI, \(Q_{AB/F}\), and VIF to 6.949, 0.724, and 0.519.

\textbf{Infrared interventions.} Zero and mean inputs increase MI to 6.537 and 6.645 while lowering SCD to 0.898 and 0.789. Spatial shuffling lowers SCD to 0.773, \(Q_{AB/F}\) to 0.569, and VIF to 0.332.

\textbf{Input-path diagnostics.} For R/B-zero and G-only inputs, all reported aggregate metrics match the normal-input results at the displayed precision, consistent with the adapter's removal of the two color-difference latents. Table~\ref{tab:infrared-controls-full} reports input copies processed through the replay pipeline. Table~\ref{tab:E3b} reports exact copies of the scoring references.

% Source: author-maintained rawprism.md; all table cells preserved.
\begin{table*}[!t]
\centering
\renewcommand{\thetable}{E2}
\renewcommand{\theHtable}{sourceE2}
\caption{Infrared-input controls from the fixed-weight replay on 280 Protocol-C scenes (Appendix~\ref{sec:E.ablations}). All rows use the Protocol-C display renderer. Model-based rows share a common reference. Input copies pass through the replay pipeline. Table~\ref{tab:E3b} gives exact copies of the scoring references. \(\mathrm{PSNR}_{\mathrm{src}}\) is source-averaged PSNR in dB. Bold/underline: highest/second-highest values (ties included).}
\label{tab:infrared-controls-full}
\small
\setlength{\tabcolsep}{2.5pt}
\renewcommand{\arraystretch}{1.06}
\begin{tabularx}{\linewidth}{@{}Xrrrrrrrrr@{}}
\toprule
Input or output control & EN & SD & \(Q_{AB/F}\) & MI & SCD & VIF & CC & SSIM & \(\mathrm{PSNR}_{\mathrm{src}}\) \\
\midrule
Normal infrared input & 7.374 & 0.218 & 0.716 & 5.187 & \textbf{1.389} & 0.489 & \textbf{0.504} & \textbf{0.704} & 19.520 \\
Infrared zero & 7.356 & 0.221 & \underline{0.724} & 6.537 & \underline{0.898} & 0.514 & \underline{0.476} & 0.702 & 23.033 \\
Infrared mean & 7.342 & 0.221 & \underline{0.724} & 6.645 & 0.789 & 0.512 & 0.474 & 0.701 & \underline{23.083} \\
Infrared spatial shuffle & \underline{7.376} & 0.216 & 0.569 & 4.188 & 0.773 & 0.332 & 0.470 & 0.597 & 19.320 \\
Infrared R/B sub-lattices zeroed & 7.374 & 0.218 & 0.716 & 5.187 & \textbf{1.389} & 0.489 & \textbf{0.504} & \textbf{0.704} & 19.520 \\
Infrared G sub-lattices only & 7.374 & 0.218 & 0.716 & 5.187 & \textbf{1.389} & 0.489 & \textbf{0.504} & \textbf{0.704} & 19.520 \\
Pipeline copy of visible input & 7.341 & \underline{0.222} & \textbf{0.725} & \textbf{7.136} & 0.710 & \underline{0.522} & 0.474 & \underline{0.703} & \textbf{23.425} \\
Pipeline copy of infrared input & \textbf{7.449} & \textbf{0.235} & 0.241 & \underline{6.717} & 0.622 & \textbf{0.563} & 0.453 & 0.656 & 13.139 \\
\bottomrule
\end{tabularx}
\end{table*}

% Source controls retained for the evaluated model and declared scoring conditions.

\subsection{Source-Copy and Averaging Controls}
\label{sec:E.3}

\textbf{Source copies and simple averaging.} The no-training MEF/MFF controls use 42 and 15 scenes, respectively. MEF includes copies of the rendered under/over inputs, their average, and an average of packed RAW inputs passed through the same DNG rendering path. All MEF controls use the same rendered under/over scoring references. The over-exposed source copy has the highest MEF MI and SCD in Table~\ref{tab:E3a}.

% Source: rawprism.md, Table~\ref{tab:E3a}; caption and cell text are preserved.
\begin{table}[!t]
\centering
\renewcommand{\thetable}{E3a}
\renewcommand{\theHtable}{sourceE3a}
\caption{MEF and MFF source-copy and averaging controls. MFF retains the legacy rendering condition. Bold/underline: highest/second-highest values (ties included). Rankings are computed separately for each task.}
\label{tab:E3a}
\small
\setlength{\tabcolsep}{3pt}
\renewcommand{\arraystretch}{1.06}
\begin{tabularx}{\linewidth}{@{}lXrr@{}}
\toprule
Task & Output & MI & SCD \\
\midrule
MEF & Under-exposed source & \underline{8.207} & −0.751 \\
MEF & Over-exposed source & \textbf{9.698} & \textbf{0.972} \\
MEF & Under/over rendered average & 8.113 & 0.221 \\
MEF & Packed-RAW linear average & 8.125 & 0.264 \\
MEF & RAWPrism final & 7.248 & \underline{0.380} \\
MFF & Far-focus source & \textbf{8.599} & 0.309 \\
MFF & Near-focus source & \underline{8.478} & 0.035 \\
MFF & Far/near average & 5.682 & \underline{0.344} \\
MFF & RAWPrism, legacy rendering & 5.758 & \textbf{0.424} \\
\bottomrule
\end{tabularx}
\end{table}

For IVIF, RAWPrism has higher SCD than both exact scoring-source copies, and both copies have higher MI (Table~\ref{tab:E3b}). Each exact copy yields infinite source-averaged PSNR because its error against the copied source is zero. The controls use the scene IDs and recorded Protocol-C condition of the main result. A complete input-hash chain linking the two runs is unavailable.

% Source: rawprism.md, Table~\ref{tab:E3b}; table cells are preserved.
\begin{table*}[!t]
\centering
\renewcommand{\thetable}{E3b}
\renewcommand{\theHtable}{sourceE3b}
\caption{Exact-source IVIF copies under the archived Protocol-C condition, 280 scenes. \(\mathrm{PSNR}_{\mathrm{src}}\) is source-averaged PSNR in dB. Bold/underline: highest/second-highest values (ties included).}
\label{tab:E3b}
\small
\setlength{\tabcolsep}{2.5pt}
\renewcommand{\arraystretch}{1.06}
\begin{tabularx}{\linewidth}{@{}Xrrrrrrrrr@{}}
\toprule
Output & EN & SD & \(Q_{AB/F}\) & MI & SCD & VIF & CC & SSIM & \(\mathrm{PSNR}_{\mathrm{src}}\) \\
\midrule
Copy visible & \underline{7.184} & \underline{0.206} & \textbf{0.752} & \textbf{8.204} & \underline{0.723} & \textbf{0.524} & \underline{0.477} & \textbf{0.722} & \(\boldsymbol{+\infty}\) \\
Copy infrared & 6.874 & 0.174 & 0.285 & \underline{7.897} & 0.677 & \underline{0.512} & \underline{0.477} & \textbf{0.722} & \(\boldsymbol{+\infty}\) \\
RAWPrism final C & \textbf{7.392} & \textbf{0.222} & \underline{0.711} & 5.284 & \textbf{1.357} & 0.495 & \textbf{0.501} & \underline{0.699} & \underline{19.081} \\
\bottomrule
\end{tabularx}
\end{table*}

\subsection{Rendering and Reference Sensitivity}
\label{sec:E.4}

\textbf{Input and reference protocols.} The IVIF comparison between Protocols C and S changes both the visible scoring reference and the RGB-baseline input. The final merged model obtains MI/SCD/\(Q_{AB/F}\)/VIF of 5.284/1.357/0.711/0.495 under C and 3.492/0.817/0.449/0.312 under S.

For MFF, Table~\ref{tab:4a} uses the legacy renderer. Table~\ref{tab:E4} reports all seven methods on 15 scenes with source white-balance metadata and corrected channel order. We resize each source globally to the fused output size and compute VIF and SSIM on the maximum-side-1024 grid.

% Source: author-maintained rawprism.md; all table cells preserved.
\begin{table*}[!t]
\centering
\renewcommand{\thetable}{E4}
\renewcommand{\theHtable}{sourceE4}
\caption{Corrected-source MFF comparison on 15 scenes. Bold/underline: highest/second-highest values (ties included).}
\label{tab:E4}
\small
\setlength{\tabcolsep}{2.5pt}
\renewcommand{\arraystretch}{1.06}
\begin{tabularx}{\linewidth}{@{}Xrrrrrrr@{}}
\toprule
Method, corrected-source MFF & EN & SD & \(Q_{AB/F}\) & MI & SCD & VIF & SSIM \\
\midrule
UniFusion & 6.647 & 0.117 & \underline{0.586} & 5.588 & 0.256 & \underline{0.646} & 0.912 \\
SwinFusion & 6.679 & 0.120 & 0.515 & \textbf{5.633} & 0.488 & 0.624 & \textbf{0.923} \\
CDDFuse & 6.303 & \textbf{0.179} & 0.330 & 3.021 & 0.557 & 0.430 & 0.530 \\
IFCNN & \underline{6.711} & 0.124 & 0.559 & \underline{5.616} & \underline{0.632} & 0.639 & \underline{0.913} \\
Laplacian & \textbf{6.758} & \underline{0.125} & \textbf{0.604} & 5.406 & \textbf{0.694} & \textbf{0.702} & 0.898 \\
RAWPrism & 6.681 & 0.123 & 0.415 & 5.033 & 0.472 & 0.604 & 0.888 \\
TC-MoA & 6.699 & 0.119 & 0.396 & 2.194 & −0.483 & 0.342 & 0.755 \\
\bottomrule
\end{tabularx}
\end{table*}

The archived records specify the corrected-source MFF and standard-protocol IVIF comparisons. Their linkage to runs with the newer opt-in fixed renderer remains unverified.

\subsection{Paired Task Comparisons}
\label{sec:E.6}

We compare methods on common scenes under the same rendering and scoring conditions. The paired-bootstrap intervals describe uncertainty in mean differences across scenes for the evaluated checkpoints (Appendix~\ref{sec:C.3}).

\subsubsection{MEF paired comparisons}

RAWPrism achieves the highest MI, SCD, and CC among the five MEF methods in Table~\ref{tab:3}. A paired bootstrap over the 42 test scenes gives an MI difference of +1.788 with interval {[}1.482, 2.058{]} relative to SwinFusion and +0.899 {[}0.601, 1.204{]} relative to Mertens. RAWPrism has higher MI on 40 and 35 scenes, respectively. The corresponding SCD differences are +0.197 {[}0.144, 0.256{]} and +0.164 {[}0.119, 0.215{]}. Paired differences use unrounded per-scene values.

Mertens has higher VIF, 1.039 versus 1.005. The paired difference for RAWPrism is −0.034 {[}−0.070, 0.008{]}. SwinFusion has the highest \(Q_{AB/F}\), while Mertens has the highest EN, SD, and MEF-SSIM.

\subsubsection{MFF rendering conditions}

The primary MFF comparison uses the legacy R/B-swapped renderer (Table~\ref{tab:4a}), and Table~\ref{tab:E4} reports results with corrected-source rendering. Under legacy rendering, RAWPrism has the highest MI, with a paired margin of +0.102 over SwinFusion and an interval {[}−0.17, 0.38{]} that includes zero. The Laplacian reference leads EN, \(Q_{AB/F}\), and VIF, and CDDFuse leads SD and SCD. RAWPrism ranks third on \(Q_{AB/F}\).

\subsubsection{IVIF metric trade-offs}

On the 280-scene FMB-derived test partition under Protocol C, RAWPrism obtains the highest EN, SD, MI, and VIF among the methods in Table~\ref{tab:5}. Its MI is 5.284, compared with 4.548 for SwinFusion, 4.352 for UniFusion, and 3.884 for TC-MoA. The paired MI difference from SwinFusion is +0.736 with interval {[}0.647, 0.823{]}. Its \(Q_{AB/F}\) of 0.711 ranks behind UniFusion at 0.716 and the Laplacian reference at 0.715.

CDDFuse and SeAFusion have the highest and second-highest SCD, respectively, and IFCNN is second on CC and SSIM.

\onecolumn
\section{Additional Qualitative Comparisons}
\label{sec:F}
\setcounter{figure}{0}
\renewcommand{\thefigure}{F\arabic{figure}}
\renewcommand{\theHfigure}{supplement.F\arabic{figure}}
\captionsetup{hypcap=false}

\noindent\begin{minipage}{\linewidth}
  \centering
  \includegraphics[width=\linewidth]{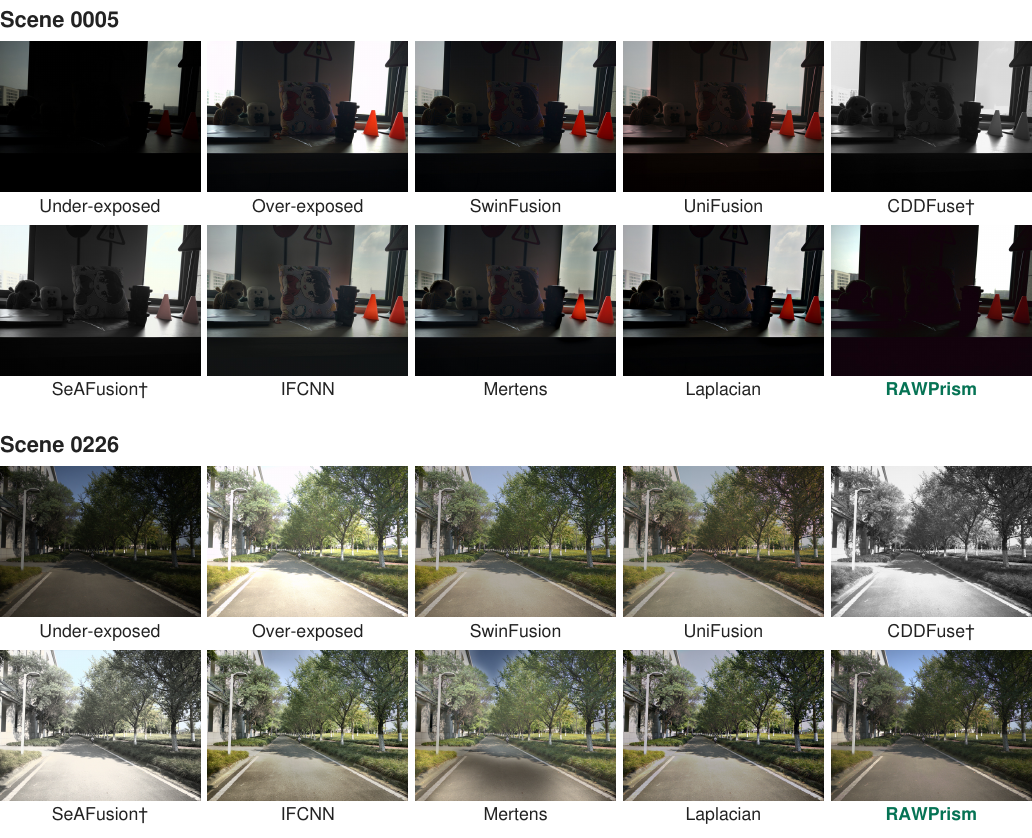}
  \captionof{figure}{Multi-exposure fusion on two additional test scenes. Each scene shows the under- and over-exposed inputs followed by outputs from eight methods. The dagger marks cross-task applications of CDDFuse and SeAFusion.}
  \label{fig:supp_qual_mef}
\end{minipage}

\clearpage
\noindent\begin{minipage}{\linewidth}
  \centering
  \includegraphics[width=\linewidth,height=.88\textheight,keepaspectratio]{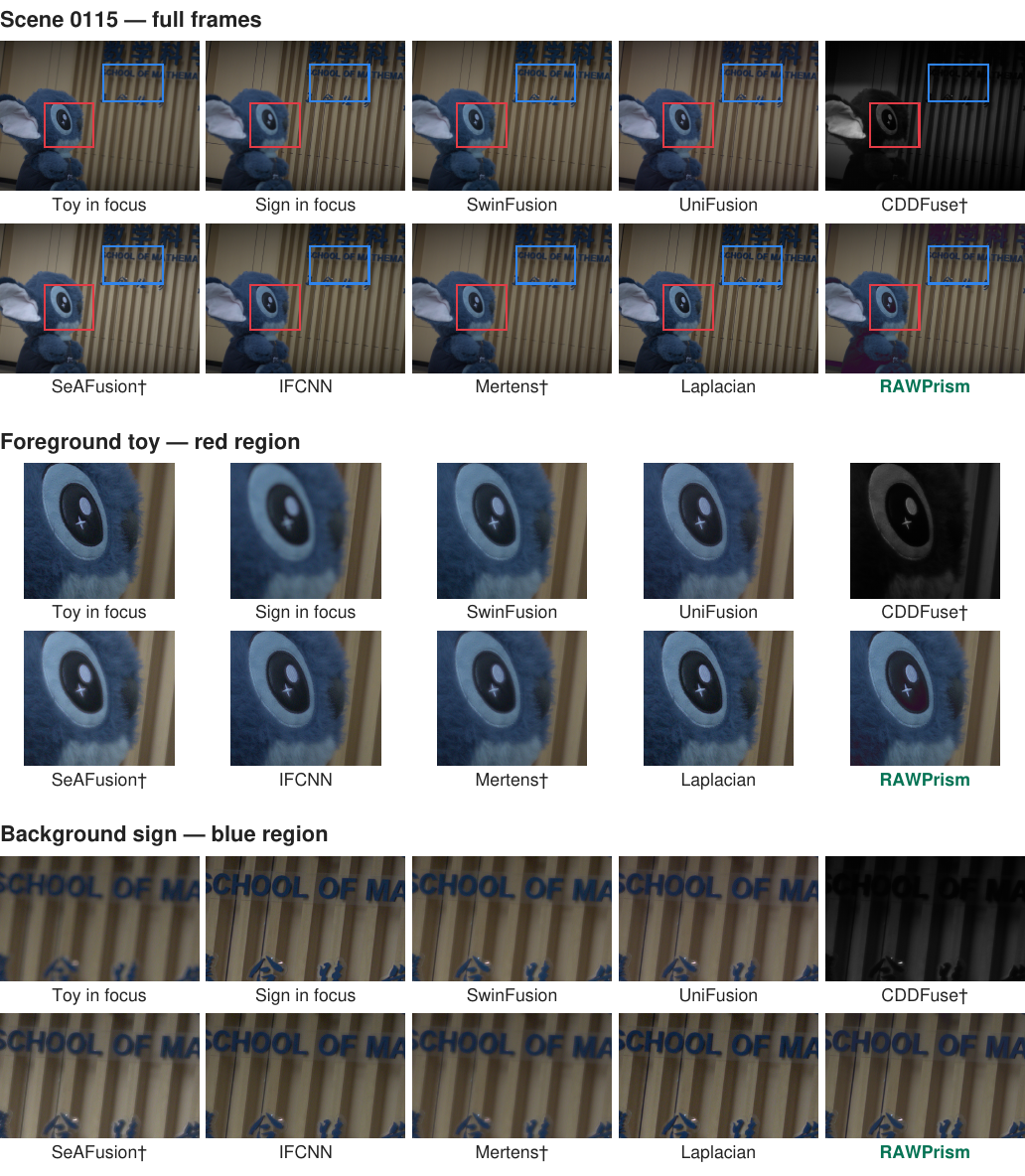}
  \captionof{figure}{Multi-focus fusion on test scene 0115, also shown in the main paper. The two inputs and all eight outputs use the same crop coordinates for the foreground toy and background sign. The two inputs and RAWPrism output use the shared metadata-based display ISP. The dagger marks cross-task applications of CDDFuse, SeAFusion, and Mertens.}
  \label{fig:supp_qual_mff}
\end{minipage}

\clearpage
\noindent\begin{minipage}{\linewidth}
  \centering
  \includegraphics[width=\linewidth,height=.88\textheight,keepaspectratio]{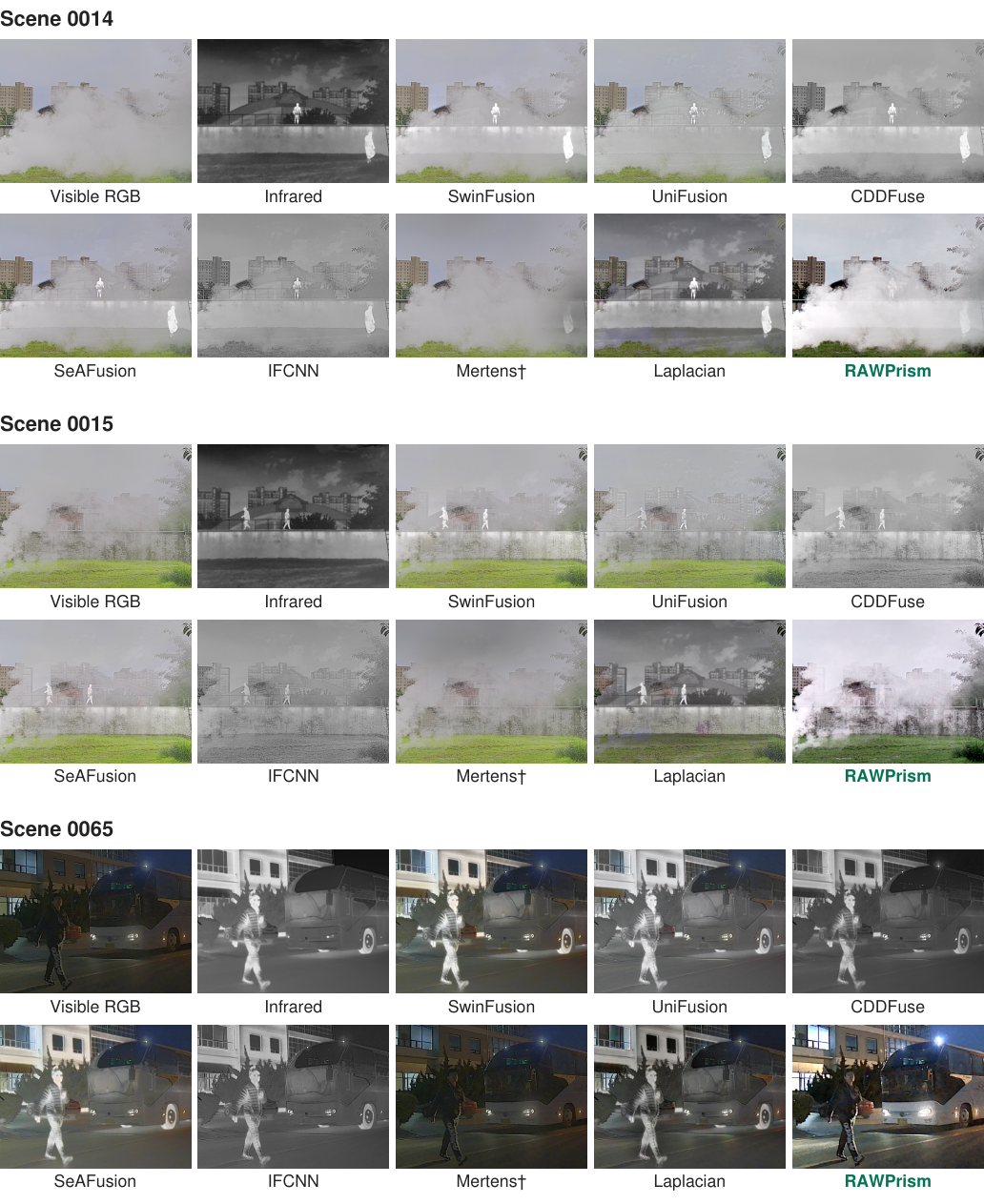}
  \captionof{figure}{Infrared--visible fusion on three additional FMB~\cite{ref1} test scenes under Protocol S. RGB baselines use the original visible RGB, and RAWPrism uses reconstructed visible RAW. Both use the same infrared observation. The dagger marks the cross-task application of Mertens.}
  \label{fig:supp_qual_ivif}
\end{minipage}

\clearpage
\section{Additional Sony \texorpdfstring{$\alpha$7~III}{alpha7 III} Captures}
\label{sec:G}
\setcounter{figure}{0}
\renewcommand{\thefigure}{G\arabic{figure}}
\renewcommand{\theHfigure}{supplement.G\arabic{figure}}

We also captured multi-exposure and multi-focus scenes with a Sony $\alpha$7~III (ILCE-7M3). The archive groups native ARW files and extracted data by task and batch. We count unique scene IDs in the extracted core batches. The paper's quantitative results use the splits in Appendix~\ref{sec:A}.

\subsection{Multi-Exposure Captures}
\label{sec:G:mef}

The two core MEF batches contain 69 and 82 scene directories, for a total of 151. The two examples in Figure~\ref{fig:sony_mef} have $6000\times4000$ default RAW crops with an RGGB mosaic and three bracket settings ($-4$, $-1$, and $+2$ EV). The exposure times for scene 0803\_1 are 1/160, 1/30, and 1/30 seconds, a longest-to-shortest ratio of $5.3\times$. Its last two frames use ISO 160 and ISO 1250, respectively. Scene 0804\_1 uses 1/4000, 1/500, and 1/60 seconds at ISO 100, a ratio of $66.7\times$.

\noindent\begin{minipage}{\linewidth}
  \centering
  \includegraphics[width=\linewidth,height=.65\textheight,keepaspectratio]{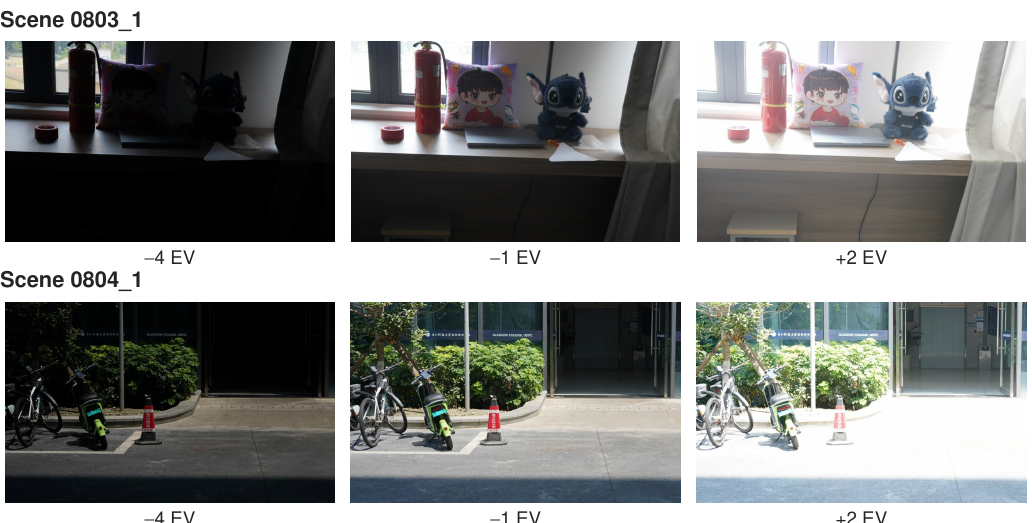}
  \captionof{figure}{Multi-exposure examples from the two Sony capture batches. Each row shows three RGB previews cropped to the same region, with EV labels from the source archive.}
  \label{fig:sony_mef}
\end{minipage}

\clearpage
\subsection{Multi-Focus Captures}
\label{sec:G:mff}

The three core MFF batches contain 23, 8, and 16 scene directories, for a total of 47. A separate test capture batch contains 12 ARW files. The scenes in Figure~\ref{fig:sony_mff} have RGGB mosaics with default RAW crops of $6000\times4000$ (0806\_1) and $3936\times2624$ (0808\_1).

The Sony RAW preprocessing pipeline compensates for focus breathing by estimating a global transform from the green CFA planes. It warps the near-focus mosaic in its packed CFA representation into the far-focus frame and crops both inputs to a common valid region. The registered pair records the transform, crop coordinates, and validity mask.

\noindent\begin{minipage}{\linewidth}
  \centering
  \includegraphics[width=\linewidth,height=.70\textheight,keepaspectratio]{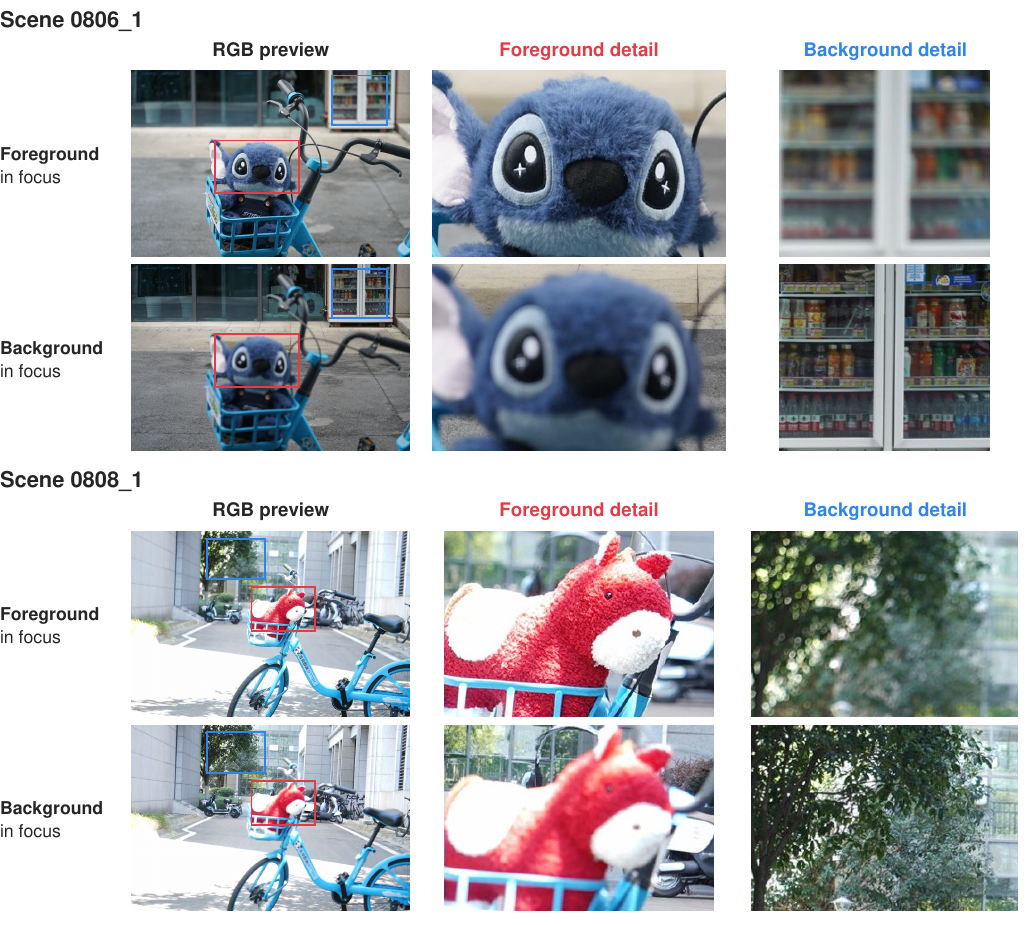}
  \captionof{figure}{Two Sony multi-focus scenes. Each scene shows the foreground- and background-focused captures in separate rows. Boxes identify the enlarged foreground and background regions in each input.}
  \label{fig:sony_mff}
\end{minipage}

\end{document}